\documentclass[10pt,twocolumn,letterpaper]{article}

\usepackage[pagenumbers]{cvpr} 

\usepackage{graphicx}
\usepackage{amsmath}
\usepackage{amssymb}
\usepackage{booktabs}
\usepackage{stfloats}
\usepackage[colorlinks,
            linkcolor=red,
            anchorcolor=blue,
            citecolor=green
            ]{hyperref}

\usepackage{booktabs}
\usepackage{colortbl}
\usepackage{multirow}
\usepackage{makecell}
\usepackage[table]{xcolor}

\newcommand{\enumroman}[1]{\textbf{\uppercase\expandafter{\romannumeral#1}}. }{}
\newcommand{\numroman}[1]{\uppercase\expandafter{\romannumeral#1} }{}

\usepackage[capitalize]{cleveref}
\crefname{section}{Sec.}{Secs.}
\Crefname{section}{Section}{Sections}
\Crefname{table}{Table}{Tables}
\crefname{table}{Tab.}{Tabs.}

\def\cvprPaperID{1536} 
\def\confName{CVPR}
\def\confYear{2023}

\begin{document}

\title{PyramidFlow: High-Resolution Defect Contrastive Localization\\
using Pyramid Normalizing Flow}
\author{
Jiarui Lei\textsuperscript{1,2}\qquad Xiaobo Hu\textsuperscript{1,2}\qquad Yue Wang\textsuperscript{1,2}\qquad Dong Liu\textsuperscript{1,2,3,4*} \\
{\fontsize{10pt}{\baselineskip}\selectfont \textsuperscript{1}State Key Laboratory of Modern Optical Instrumentation, Zhejiang University.}\\
{\fontsize{10pt}{\baselineskip}\selectfont \textsuperscript{2}ZJU-Hangzhou Global Scientific and Technological Innovation Center, China}\\
{\fontsize{10pt}{\baselineskip}\selectfont \textsuperscript{3}Jiaxing Key Laboratory of Photonic Sensing \& Intelligent Imaging, China}\\
{\fontsize{10pt}{\baselineskip}\selectfont \textsuperscript{4}Intelligent Optics \& Photonics Research Center, Jiaxing Research Institute Zhejiang University}\\
{\tt\small \{karrilett, huxiaobo, 426195, liudongopt\}@zju.edu.cn}
}

\maketitle

\begin{abstract}

During industrial processing, unforeseen defects may arise in products due to uncontrollable factors. 
Although unsupervised methods have been successful in defect localization, the usual use of pre-trained models results in low-resolution outputs, which damages visual performance.
To address this issue, we propose PyramidFlow, the first fully normalizing flow method without pre-trained models that enables high-resolution defect localization. 
Specifically, we propose a latent template-based defect contrastive localization paradigm to reduce intra-class variance, as the pre-trained models do. In addition, PyramidFlow utilizes pyramid-like normalizing flows for multi-scale fusing and volume normalization to help generalization.
Our comprehensive studies on MVTecAD demonstrate the proposed method outperforms the comparable algorithms that do not use external priors, even achieving state-of-the-art performance in more challenging BTAD scenarios.

\end{abstract}


\section{Introduction}
\label{sec:intro}

Due to the uncontrollable factors in the complex industrial manufacturing process, unforeseen defects will be brought to products inevitably.
As the human visual system has the inherent ability to perceive anomalies\cite{TIM-Survey}, quality control relies on manual inspection for a long time. 

However, large-scale images and tiny defects are challenging for manual inspection, so increasing research is focused on automated machine vision inspection.
Among all the methods, supervised deep learning has achieved great success.
It relies on annotated datasets to learn discriminative features, effectively overcoming the hand-crafted shortcomings. 
However, because of insufficient negative samples, the high demand for labels, and the absence of prior knowledge, those approaches based on supervised learning may suffer in identifying unseen defects in practices, 

\begin{figure}
    \centering
    \setlength{\abovecaptionskip}{0.cm} 
    \setlength{\belowcaptionskip}{-0.8cm} 
    \includegraphics[width=0.9\linewidth]{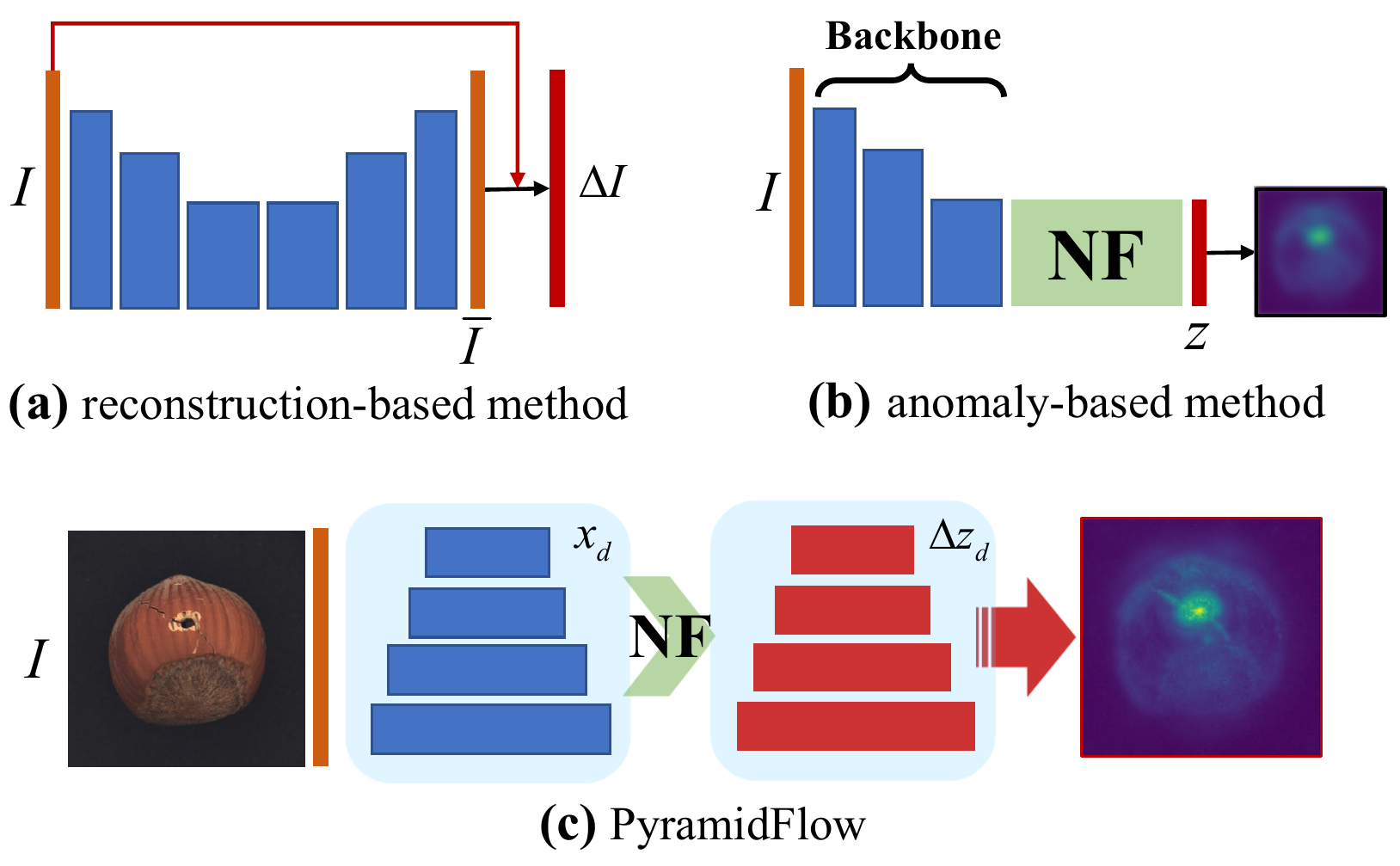}
    \caption{
        Illustration of various anomaly localization methods.
        \textbf{(a)} Reconstruction-based method.
        \textbf{(b)} Anomaly-based method, where {\bf NF} denotes normalizing flow.
        \textbf{(c)} Our PyramidFlow, which combines latent templates and normalizing flow, enables high-resolution localization.
    }
    \label{fig1}
\end{figure}

Recently, unsupervised methods have been applied to defect detection, as shown in \cref{fig1}(a,b). 
Reconstruction-based methods\cite{AE-SSIM,VanillaVAE,AnoGAN,DRAEM} are the most famous, which take reconstructed images as templates and then apply explicit contrast in image space to achieve high-resolution localization. However, reconstructing using decoders is an ill-posed inverse problem, it is hard to reconstruct complex details.
To overcome the above limitations, anomaly-based methods\cite{PaDiM,SPADE} utilizing texture-aware pre-trained models achieves high image-level performance, which also damages pixel-level visual performance. One of the most promising methods is convolutional normalizing flows\cite{FastFlow,CFlow,CSFlow}, which models the probability distribution further from pre-trained features, earning higher performance.

In this paper, a Pyramid Normalizing Flow (PyramidFlow) is proposed. It develops the idea of templates from image space into latent space by normalizing flow, then performing contrast $\Delta z_d$ for high-resolution anomaly localization, as shown in \cref{fig1}(c). 
Specifically, we propose the multi-scale Pyramid Coupling Block, which includes invertible pyramid and volume normalization, as the critical module to construct volume-preserving PyramidFlow.
To the best of our knowledge, PyramidFlow is the first UNet-like fully normalizing flow specifically designed for anomaly localization, analogous to UNet\cite{UNet} for biomedical image segmentation. Our main contributions can be summarized as follows.

\begin{itemize}
\setlength{\itemsep}{0.5em}
\setlength{\parsep}{0pt}
\setlength{\parskip}{0pt}
    \item We propose a latent template-based defect contrastive localization paradigm. Similar to the reconstruction-based methods, we perform contrast localization in latent space, which avoids the ill-posedness and reduces intra-classes variance efficiently.
    
    \item We propose PyramidFlow, which includes invertible pyramids and pyramid coupling blocks for multi-scale fusing and mapping, enabling high-resolution defect localization. Additionally, we propose volume normalization for improving generalization. 
    
    \item We conduct comprehensive experiments to demonstrate that our advanced method outperforms comparable algorithms that do not use external priors, and even achieves state-of-the-art performance in complex scenarios.
   
\end{itemize}

\section{Related Work}
\label{sec:related}

\subsection{Deep learning-based Defect Localization}

With the rise of deep learning, numerous works apply generalized computer vision methods for defect detection. 
Some works are based on object detection\cite{FPCB,CADN,FusionDetect,ES-Net,DefectNet}, which relies on annotated rectangular boxes, enabling locating and classifying defects end-to-end. 
The other is applying semantic segmentation\cite{SurfaceInspection,SurfaceDetect,Mixed-SDD}, which enables pixel-level localization, suit for complex scenarios with difficult-to-locate boundaries. However, these works still rely on supervised learning, they attempt to collect sufficient defective samples to learn well-defined representations.

Recently, some promising work has considered the scarcity of defects in real-world scenarios, where defect-free samples are only obtained. 
These methods can be classified as reconstruction-based and anomaly-based. 
The reconstruction-based method relies on generative models such as VAE or GAN, which encode  a defective image and reconstruct it to a defect-free image, then localize the defect with the contrast of these two images.
The reconstruction-based method performs well on single textural images, but they cannot generalize to non-textural images for ill-posedness and degeneracy\cite{TIM-Survey}.
The anomaly-based method treats defects as anomalous, applying neural networks to discriminate between normality and anomalous.
These methods extract pre-trained features, then estimate their probability density using Mahalanobis distances or K-NearestNeighbor, while the lower probability indicates where the image patches are abnormal.
Although anomaly-based methods had achieved great success in defect detection, it locates defects with low pixel-level resolution compared with reconstruction-based methods, usually 1/16th or even lower, which greatly limits practical industrial applications.

To overcome existing shortcomings, we propose a latent template-based defect contrastive localization paradigm, which breaks the limitation of low-frequency texture-aware-only models, enabling more accurate results. 

\subsection{Normalizing Flow}

Normalizing flow is a kind of invertible neural network with bijective mappings and traceable Jacobi determinants.
It was first proposed for nonlinear independent component estimation\cite{NICE} and applied to anomaly detection\cite{DifferNet} recently for its invertibility helps prevent mode collapse. 
The normalizing flow comprises coupling blocks, these basic modules for realizing nonlinear mappings and calculating Jacobi determinants. Originally, NICE\cite{NICE} proposed the additive coupling layer with unitary Jacobi determinants, while RealNVP\cite{RealNVP} further proposed the affine coupling layer that enables the generation of non-volume-preserving mappings. 
However, redundant volume degrees of freedom can lead to increased optimization complexity, creating a domain gap between maximum likelihood estimation and anomaly metrics, which may potentially compromise the generalization performance in anomaly detection. 

Previous works\cite{DifferNet,CFlow,FastFlow} on anomaly localization usually follow the methods proposed in RealNVP, but some challenges remain. Some studies\cite{CSFlow} have found that convolutional normalizing flow focuses on local rather than semantic correlations, which are usually addressed by image embeddings\cite{WhyNFFail}. Hence, earlier studies\cite{DifferNet} adopted pre-trained backbones, while recent trends used pre-trained encoders to extract image patches\cite{CSFlow,CFlow,FastFlow}. However, pre-trained-based methods rely on task-irrelevant external priors, which limit generalization in unforeseen scenarios. 

To address the above challenges, we propose a pyramid-like normalizing flow called PyramidFlow, which utilizes volume normalization to preserve volume mappings that include task-relevant implicit priors. Additionally, our method offers the option of using pre-trained models, and we have observed that external priors from pre-trained models can improve generalization performance. We will discuss these contributions in \cref{Sec43} and \cref{Sec44}.

\section{Methodology}
\label{Sec3}

\begin{figure*}
    \centering
    \setlength{\abovecaptionskip}{0.cm} 
    \setlength{\belowcaptionskip}{-0.7cm} 
    \includegraphics[width=0.95\linewidth]{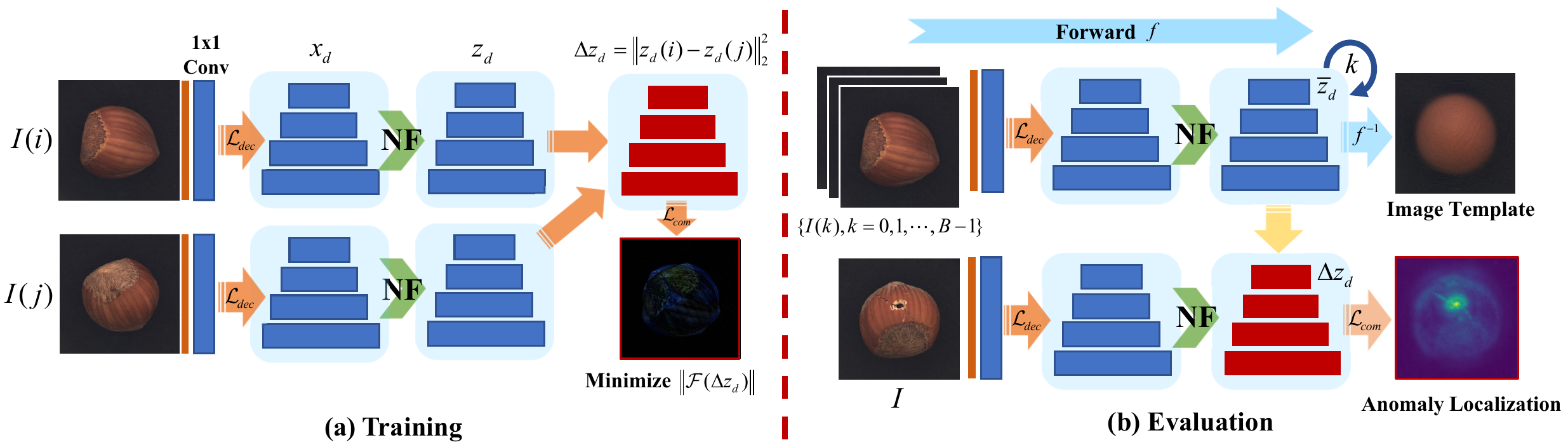}
    \caption{
        Schematic of training and evaluation for PyramidFlow.
        \textbf{(a)} Given any normality pair, minimize the distance of latent variables.
        \textbf{(b)} The means of the latent variables are contrasted to the examples, then apply pyramid composition to obtain an anomaly localization map.
    }
    \label{fig2}
\end{figure*}

Our algorithm consists of two processes, training and evaluation, as shown in \cref{fig2}.
The training process is similar to siamese networks, the model is optimized by minimizing the Frequency differences $\|\mathcal{F}(\Delta z_d)\|$ within the image pair.
For the evaluation process, latent templates are obtained through inference at the total training dataset, then latent contrast and pyramid composition are applied to obtain an anomaly localization map. The details are shown in the following sections.

\subsection{Invertible Pyramid}\label{Sec31}
Defect images contain various frequency components. Usually, the low-frequency components represent the slow gradient background, while the high-frequency components correspond to details or defects.
To decouple the frequency components and identify each frequency component independently, we propose invertible pyramids, which enable multi-scale decomposition and composition for a single feature.
To facilitate feature learning, previous work applies pre-trained encoders to extract features. Although pre-trained methods with external priors help performance improvement, to fully explore the advantages of our approach in our primary study, let's consider a baseline without any pre-trained model.

For a three-channel image $I$, apply orthogonally initialized $1\times 1$ convolution $\mathbf{W}\in \mathbb{R}^{C\times 3}$ to the image for obtain features $x=\mathbf{W}I$. Given a feature $x$ and a positive integer $L$, the pyramid decomposition is a mapping from features to feature sets $\mathcal{L}_{dec}:x\to \{x_d| d\in \mathbb{Z}_{L-1}\}$, where $x_d$ is the $d-$level pyramid can be calculated as

\begin{equation}
    x_d = D^d(x) - U(D^{d+1}(x))
    \label{eq1:decomposition}
\end{equation}

\noindent where $D(\cdot)$ and $U(\cdot)$ are arbitrary linear upsampling and downsampling operators, while $D^d(\cdot)$ represents repeated downsampling $d$ times. If \cref{eq1:decomposition} is further satisfied $D^0(x)=x, D^L(x)=0$, then the inverse operation $\mathcal{L}_{com}:\{x_d| d\in \mathbb{Z}_{L-1}\}$ of the pyramid decomposition can be described as

\begin{equation}
    x=\sum_{d=0}^{L-1}{U^d(x_d)}
    \label{eq2:composition}
\end{equation}

Differing from Gaussian pyramid, \cref{eq2:composition} indicates that there is always an inverse operation for pyramid decomposition, which is called pyramid composition.
The method based on \cref{eq1:decomposition,eq2:composition}, enabling perform multi-scale feature decomposition and composition, is a critical invertible module for PyramidFlow.

\subsection{Pyramid Coupling Block}\label{Sec32}
\noindent\textbf{Invertible Modules.}
Invertible modules are the essential elements to implementing invertible neural networks. The invertible modules introduced in this paper are invertible convolution, invertible pyramid, and affine coupling block. The affine coupling block is the basic module that constitutes the normalizing flow. It is based on feature splitting for invertible nonlinear mappings with easily traceable Jacobian determinants and inverse operations.

\begin{figure*}
    \centering
    \setlength{\abovecaptionskip}{0.cm} 
    \setlength{\belowcaptionskip}{-0.7cm} 
	\begin{subfigure}[c]{0.55\linewidth}
		\centering
		\includegraphics[width=0.9\linewidth]{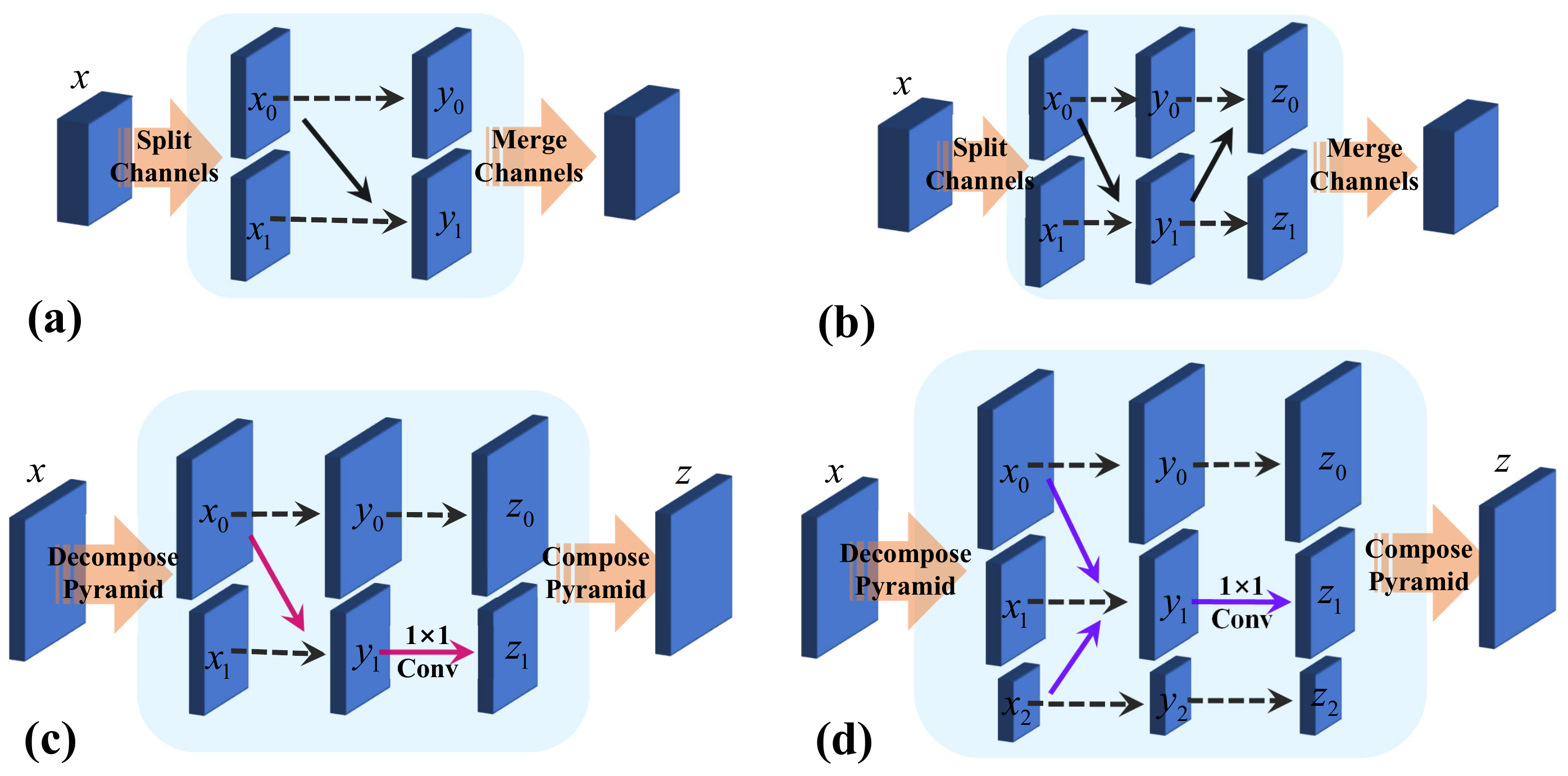}
	\end{subfigure}
	\begin{subfigure}[c]{0.4\linewidth}
		\centering
		\includegraphics[width=0.9\linewidth]{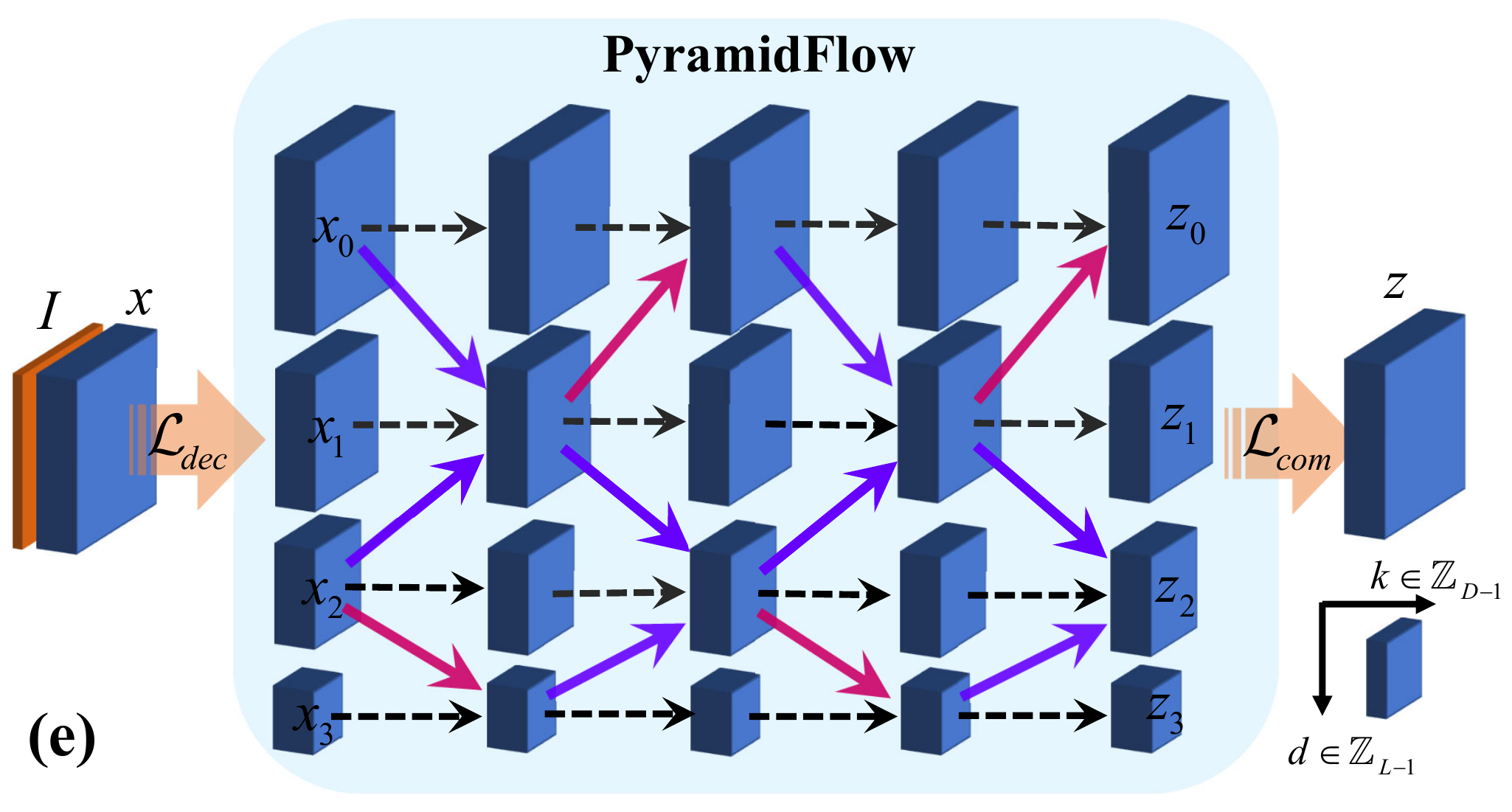}
	\end{subfigure}
	\caption{
	    The proposed pyramid coupling block and PyramidFlow, where the solid line symbolizes the transformation while the dotted line refers to identity.
	    \textbf{(a)} Channel-splitting affine coupling block.
	    \textbf{(b)} The reverse cascade of (a)-architecture.
	    \textbf{(c)} The proposed scale-wise pyramid coupling block.
	    \textbf{(d)} The reverse parallel and reparameterized of (c)-architecture.
	    \textbf{(e)} The proposed PyramidFlow, is a stacking of (c,d)-architecture both in depths and layers, where $1\times 1$ convolution is neglected to represent.
	}
    \label{fig3}
\end{figure*}

As shown in \cref{fig3}(a), the conventional affine coupling block splits a single feature along the channel dimension, where one sub-feature keeps its identity while another is performed affine transformation controlled by it.
Denote the splitted features are $x_0,x_1$ and its outputs are $y_0,y_1$, then the corresponding transformation can be described as

\begin{equation}
\begin{aligned}
    y_0&=x_0\\
    y_1&=\exp{(s(x_0))}\odot x_1+t(x_0)
    \label{eq3:affine_forward}
\end{aligned}
\end{equation}

\noindent where $s(\cdot),t(\cdot)$ are affine parameters, can be estimated by zero-initialized convolutional neural networks. For formula(\ref{eq3:affine_forward}), there is an explicit inverse transformation:

\begin{equation}
\begin{aligned}
    x_0&=y_0\\
    x_1&=\exp{(-s(y_0))}\odot (y_1-t(y_0))
    \label{eq4:affine_inverse}
\end{aligned}
\end{equation}

Denote the element at position $i,j$ of $s(\cdot)$ as $s_{i,j}(\cdot)$. As the Jacobian matrix of transformation(\ref{eq3:affine_forward}) is a triangular matrix, its logarithmic determinant can be estimated as

\begin{equation}
\begin{aligned}
    \log\left|\frac{\partial (y_0,y_1)}{\partial (x_0,x_1)}\right|=\sum_{i,j}s_{i,j}(x_0)
    \label{eq5:logdet}
\end{aligned}
\end{equation}

\cref{eq3:affine_forward,eq4:affine_inverse,eq5:logdet} are the basis of all affine coupling blocks. However, the coupling block shown in \cref{fig3}(a) remains identical for one part. Therefore the reverse cascade architecture is proposed in NICE\cite{NICE} such that both parts are transformed, as shown in  \cref{fig3}(b). The previous works construct the holistic invertible normalizing flow by iterative applying the structure shown in  \cref{fig3}(b).

\noindent\textbf{Implementation.}
Our method decomposes a single feature along the scale and realizes multi-scale feature fusion based on \cref{eq3:affine_forward,eq4:affine_inverse,eq5:logdet}. In our implementation, the multi-scale affine parameters $s(\cdot),t(\cdot)$ are estimated using a convolutional neural network with two linear layers, where bilinear interpolation is applied to match the target shape.

In addition, we employ invertible 1x1 convolution\cite{Glow} for feature fusion within features. Specifically, denoting the full rank matrix corresponding to the invertible $1\times 1$ convolution as $\mathbf{A}$, which can be decomposed by PLU as

\begin{equation}
    \mathbf{A}=\mathbf{P}\mathbf{L}(\mathbf{U} + \text{diag}(\exp (s_i)) )
    \label{eq6:PLU}
\end{equation}

\noindent where $\mathbf{P}$ is a frozen permutation matrix, $\mathbf{L}$ is a lower triangular matrix with unit diagonal elements, $\mathbf{U}$ is an upper triangular matrix with zero diagonal elements, and $\exp (s_i)$ is the $i$-th eigenvalue of the matrix $\mathbf{A}$, which always holds nonnegativity. The matrix $\mathbf{A}$ is always invertible during optimization, then its logarithmic Jacobian determinant can be estimated as

\begin{equation}
    \log |\mathbf{A}| = \sum_i s_i
    \label{eq7:Alogdet}
\end{equation}

In summary, \cref{eq3:affine_forward,eq4:affine_inverse,eq5:logdet,eq6:PLU,eq7:Alogdet} describe proposed pyramidal coupling block mathematically, as shown in \cref{fig3}(c). First, multi-scale feature fusion(\ref{eq3:affine_forward}-\ref{eq5:logdet}) is performed, and then apply linear fusion(\ref{eq6:PLU}-\ref{eq7:Alogdet}) for shuffle channels.
Furthermore, we propose a dual coupling block as shown in \cref{fig3}(d), which is equivalent to the reverse parallel of the coupling block in \cref{fig3}(c). The dual coupling block is reparameterized in our implementation, and its affine parameters $s(\cdot),t(\cdot)$ are estimated from concatenated features.

\noindent\textbf{Volume Normalization.}\label{Sec33}
Suppose that the invertible transformation $f:x\to z$ maps the variable $x$ to the latent variable $z$. Previous works have assumed that the latent variable follows basic probability distribution (\eg Gaussian distribution),  then estimates sample probability density based on the following equation:

\begin{equation}
    P(x)=P(z)\left|\frac{\partial f(x)}{\partial x}\right|
    \label{eq8:probtrans}
\end{equation}

However, this approach relies on the basic distribution assumption and ignores the effect of the implicit prior in the probability density transform on generalization. When such approaches are applied to anomaly detection, the inconsistency between the training objectives and the anomaly evaluation results in domain gaps.

\begin{figure}
    \centering
    \setlength{\abovecaptionskip}{0.cm} 
    \setlength{\belowcaptionskip}{-0.7cm} 
	\includegraphics[width=.8\linewidth]{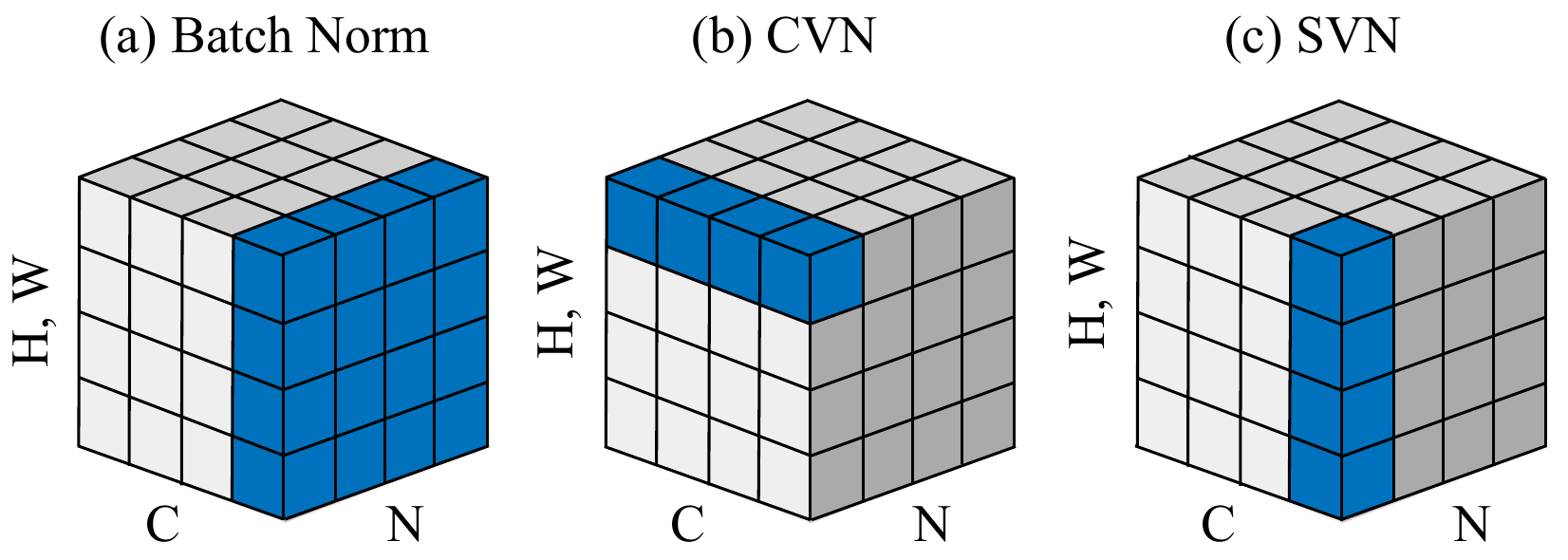}
	\caption{
	    Illustration of volume normalization.
	    \textbf{(a)} Batch Normalization.
	    \textbf{(b)} The proposed Channel Volume Normalization (CVN).
	    \textbf{(c)} The proposed Spatial Volume Normalization (SVN).
	}
    \label{fig-norm}
\end{figure}

Similar to batch normalization or instance normalization in deep learning, the proposed volume normalization will be employed for volume-preserving mappings, as illustrated in  \cref{fig-norm}.
Particularly, for the affine coupling block, the parameter $s(\cdot)$ is subtracted from its mean value before performing \cref{eq3:affine_forward}; for the invertible convolution, the parameter $s_i$ is subtracted from its mean value before calculating the matrix $\mathbf{A}$ based on \cref{eq6:PLU}. Depending on the statistical dimension, we propose Spatial Volume Normalization (SVN) and Channel Volume Normalization (CVN). SVN performs mean statistics along the spatial dimension, while CVN is along the channel dimension. Various volume normalization methods contain different priors, then we will explore their impact in \cref{Sec42}.

\subsection{Pyramid Normalizing Flow}
\noindent\textbf{Architecture.}
Our PyramidFlow can be obtained by stacking the pyramid coupled blocks of \cref{fig3}(c,d) along the depth $D-1$ times and along the layer $L-1$ times, as shown in \cref{fig3}(e). 
Specifically, PyramidFlow boosts the image $I$ to feature $x$ using matrix $\mathbf{W}$, then performs the pyramid decomposition based on \cref{eq1:decomposition}. The pyramid coupling blocks described in \cref{eq3:affine_forward,eq4:affine_inverse,eq5:logdet,eq6:PLU,eq7:Alogdet} are calculated in the order described in \cref{fig3}(e) to obtain potential pyramid features $z_d, d=0,1,\cdots, L-1$, which are finally composed into latent variables according to \cref{eq2:composition}.

\noindent\textbf{Loss Function.}
In the cases with volume normalization, the loss function excludes the probability density coefficients. Moreover, the logarithmic Jacobian determinant of the semi-orthogonal matrix $\mathbf{W}$ is sample-independent, so its effect could be ignored during training.

Suppose a training batch with 2 normal samples, its latent variables are $z_d(i), z_d(j)$.
Previous studies train neural networks using spatial difference $\Delta z_d=\|z_d(i)-z_d(j)\|^2$. However, it ignored the impact of high-frequency defects. To address the above shortcoming, we propose the following Fourier loss function.

\begin{equation}
    \mathcal{L}_{loss}=\left\|\mathcal{F}(\mathcal{L}_{com}(\{\Delta z_d|d\in \mathbb{Z}_{L-1} \}) )\right\|
    \label{eq9:fourierloss}
\end{equation}

\noindent where $\mathcal{F}$ is the fast Fourier transform of the image. Training the normalizing flow using \cref{eq9:fourierloss} enables the model to focus on the high-frequency, allowing faster convergence. We will discuss this trick in \cref{Sec43}.

\noindent\textbf{Defect Localization.}
Previous studies\cite{CSFlow,FastFlow} usually localize defects with obvious differences based on category-independent zero templates. 
In our method, the defects are modeled as anomalous deviations with respect to the template. Then, the anomaly of the latent pyramid $z_d$ is defined as $\sigma(z_d)=\|z_d-\Bar{z}_d\|$, where $\Bar{z}_d$ is mean of the latent pyramid. Finally, the total anomaly can be estimated as

\begin{equation}
    \sigma(z)=\mathcal{L}_{com}(\{\sigma(z_d)| d\in \mathbb{Z}_{L-1} \})
    \label{eq10:anomalious}
\end{equation}

The \cref{eq10:anomalious} shows that the total anomaly is a composition of anomalies at various scales, which is consistent with the empirical  method proposed by Rudolph, \etal\cite{CSFlow}.

\noindent\textbf{Image Template Estimation.}
The image template is a prototype of normal samples, a visualization of the latent template. 
Our fully normalizing flow is based on $1\times 1$ convolution instead of pre-trained encoders, maintaining end-to-end and near-invertibility, thus the flow's input $x_{temp}$ can be retrieved using \cref{eq2:composition} and \cref{eq4:affine_inverse} from latent mean $\Bar{z}_d$, then solve the least square problem $\mathbf{W}I_{temp}=x_{temp}$ for image template $I_{temp}$.

\section{Experiment and Discussion}

In this chapter, we perform unsupervised anomaly localization experiments on MVTec Anomaly Detection Dataset\cite{MVTecAD} (MVTecAD) and BeanTech Anomaly Detection Dataset\cite{VT-ADL} (BTAD). 
MVTecAD contains 15 categories of industrial defect images, five of which are textural images and the other ten are object images. The object images contain three classes (grid, metal nut, screw) without rough registration and one class (hazelnut) without fine registration, and we will discuss these cases in \cref{Sec45}.
The BTAD contains three types of real-world and industry-oriented textural images, which is more challenging for pixel-level localization.

All experiments take Area Under the Receiver Operating characteristic Curve (AUROC) and Area Under the Per Region Overlap (AUPRO) as evaluation metrics.
AUROC is the most widely used anomaly evaluation metric, and higher values indicate that various thresholds have less impact on performance. However, AUROC prefers larger anomalies and may fail in small proportions of anomalies. 
Thus, we further evaluate AUPRO for localization metric, similar to Intersection Over Union (IoU) commonly used in semantic segmentation. Detailed definitions can be found in \cite{MVTecAD}.

\subsection{Complexity Analysis}\label{Sec41}
\begin{figure}
    \centering

    \includegraphics[width=0.9\linewidth]{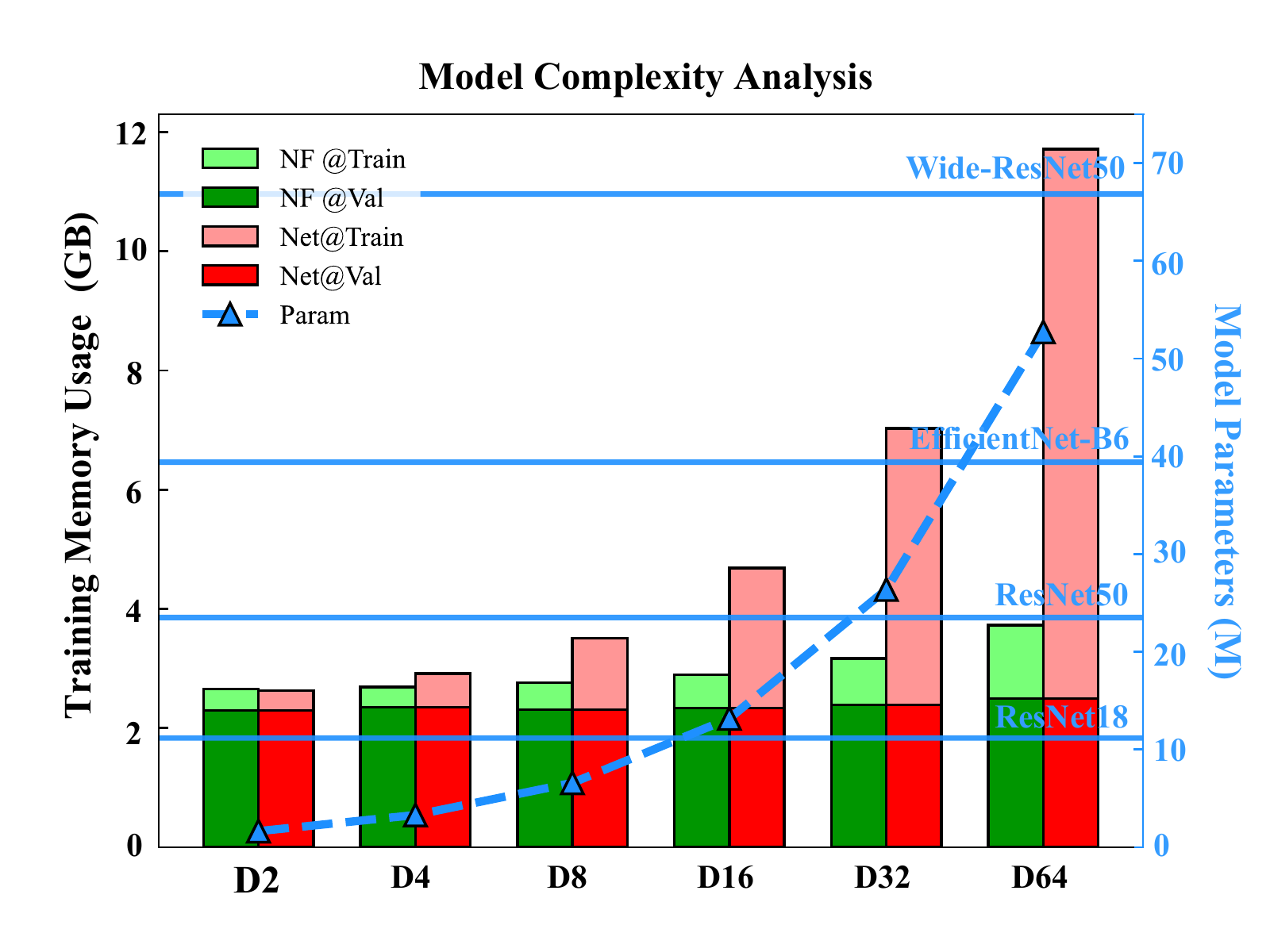}
    \caption{
        Analysis of model complexity for various depths.
        The bars correspond to the Training Memory Usage (GB) in the left vertical coordinate, while the line graph and horizontal lines relate to the Model Parameters (M) on the other side. 
        For each depth $D$, the left bar presents the normalizing flow implemented based on \textit{autoFlow} framework with memory-saving tricks, while the right is implemented by PyTorch with auto-differentiation. 
    }
    \label{fig5}
\end{figure}

The normalizing flow based on \cref{eq3:affine_forward,eq4:affine_inverse} is computationally invertible, which indicates that only one copy of the variables is necessary for all stages.
This feature decreases the memory footprint during backpropagation from linear to constant complexity. We have analyzed the above characteristics based on a fixed number of pyramid layers $L=8$, image resolution with  $256\times 256$, and channels $C=24$, then changed the number of stacked layers $D$ to explore the trends of memory usage and model parameters.
The forward and memory-saving backpropagation is implemented based on the self-developed PyTorch-based\cite{Pytorch} framework \textit{autoFlow}.
All indicators are recorded during steady-state training, then plotted as bar and line graphs, as shown in \cref{fig5}.

The memory usage based on auto-differentiation increases linearly with depth $D$, while the implementation based on normalizing flow achieves approximate depth-independent memory usage.
The memory superiority enables the proposed method to be trained in memory-constrained devices below 4G without powerful hardware.
The line graph shows the exponential trends between model parameters and depth, while the horizontal line represents the parameters of the usual pre-trained model.
We mainly adopt methods with $D<8$ or even shallower, where the number of parameters is far smaller than popular pre-training-based methods.
In summary, in scenarios of memory constraint, the proposed PyramidFlow enables dealing with larger images and requires fewer parameters than others.

\subsection{Study on Volume Normalization}\label{Sec42}
In this subsection, based on MVTecAD, we investigate the impact of volume normalization on generalization.
The experiment without data augmentation, fixed pyramid layers $L=4$, channels $C=16$, and linear interpolated image to $256\times 256$.
During the training, the volume normalization applies sample mean normalization and updates the running mean with 0.1 momenta, while in the testing, the volume normalization is based on the running mean.
We sufficiently explored the volume normalization methods proposed in \cref{Sec32}. Some representative categories are shown as \cref{tab1}.

\begin{table}[htbp]
    \centering
    \setlength{\abovecaptionskip}{0.cm} 
    \setlength{\belowcaptionskip}{-0.3cm} 
    \caption{Quantitative results of CVN and SVN on different categories. For each case in the table, the first column is \colorbox{green!5}{Pixel-AUROC\%} and the second is \colorbox{blue!5}{AUPRO\%}, while the values within parentheses represent the relative improvement.}

    \resizebox{.45\textwidth}{!}{
        
\begin{tabular}{c>{\columncolor{green!5}}c>{\columncolor{blue!5}}c>{\columncolor{green!5}}c>{\columncolor{blue!5}}c}
    \toprule
    \multirow{2}[4]{*}{\textbf{Classes}} & \multicolumn{2}{c}{\textbf{CVN}} & \multicolumn{2}{c}{\textbf{SVN}} \\
    \cmidrule{2-5}    \multicolumn{1}{c}{} & AUROC & AUPRO & AUROC & AUPRO \\
    \midrule
    capsule    & \textbf{96.1\footnotesize{(+2.6)}}  & \textbf{93.1\footnotesize{(+5.1)}} & 93.5\footnotesize{(+0.0)}  & 88.0\footnotesize{(+0.0)} \\
    pill       & \textbf{96.2\footnotesize{(+1.8)}}  & \textbf{96.3\footnotesize{(+1.4)}} & 94.4\footnotesize{(+0.0)}  & 94.9\footnotesize{(+0.0)} \\
    toothbrush & \textbf{98.9\footnotesize{(+2.5)}}  & \textbf{97.9\footnotesize{(+4.3)}} & 96.4\footnotesize{(+0.0)}  & 93.6\footnotesize{(+0.0)} \\
    carpet     &  88.9\footnotesize{(+0.0)}          & 88.3\footnotesize{(+0.0)}          & \textbf{90.8\footnotesize{(+1.9)}}  & \textbf{91.0\footnotesize{(+2.7)}} \\
    grid       &  86.2\footnotesize{(+0.0)}          & 84.5\footnotesize{(+0.0)}          & \textbf{94.2\footnotesize{(+8.0)}}   & \textbf{92.7\footnotesize{(+8.2)}} \\
    zipper     &  92.2\footnotesize{(+0.0)}          & 91.9\footnotesize{(+0.0)}          & \textbf{95.4\footnotesize{(+3.2)}}   & \textbf{95.1\footnotesize{(+3.2)}} \\
    \bottomrule
\end{tabular}%

    }

    \label{tab1}%
\end{table}%
\vspace{-0.5em}

\begin{figure}
    \centering
    \setlength{\abovecaptionskip}{0.cm} 
    \setlength{\belowcaptionskip}{-0.6cm} 
    
    \includegraphics[width=0.8\linewidth]{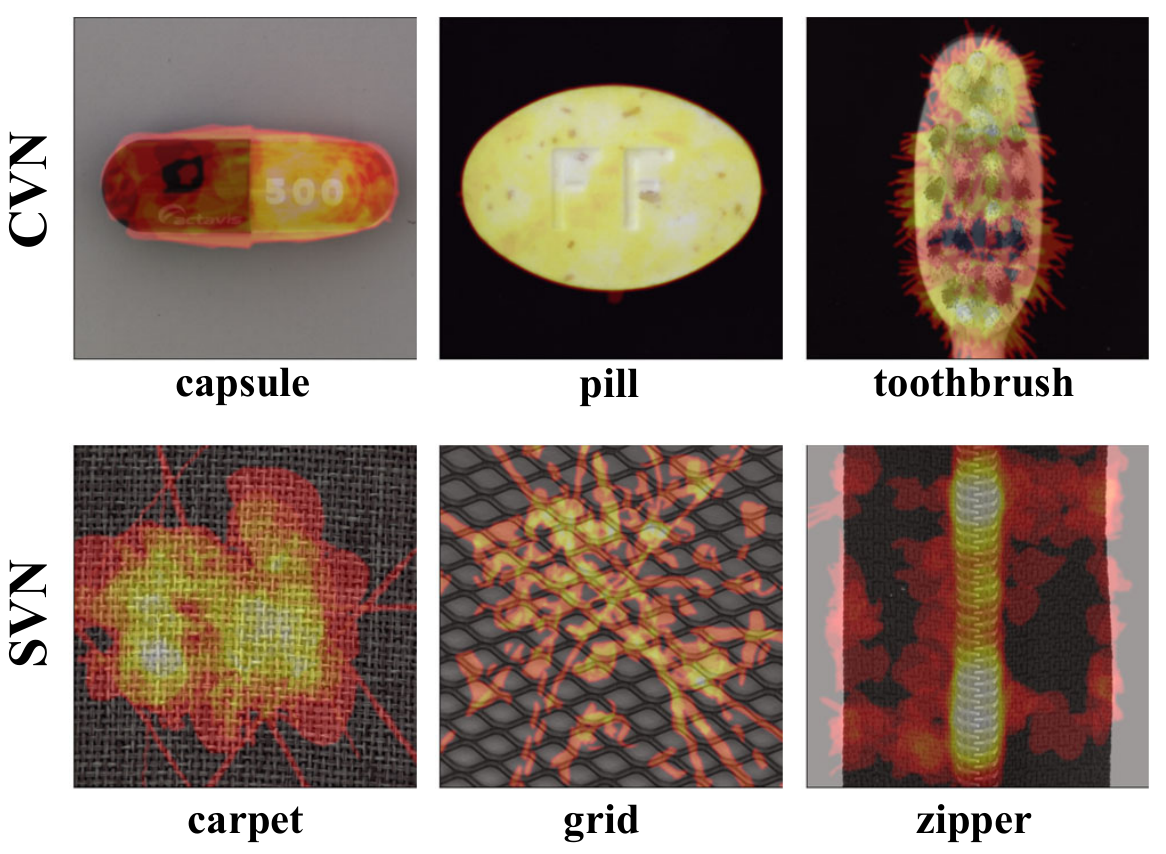}
    
    \caption{
        Defects in \cref{tab1} are visualized as heat maps. The top row displays CVN-superior class object images, while the bottom row displays SVN-superior class texture images.
    }
    \label{fig4}
\end{figure}

The result in \cref{tab1} shows the performance differences between the various volume normalization methods: CVN outperforms SVN for the first three classes, while the latter behaves the opposite. We further visualize these defect distributions in \cref{fig4}, which shows that SVN-superior classes are commonly textural images with a larger range of defects, while CVN-superior classes are object images. 

SVN with larger receptive fields achieves non-local localization by aggregating an extensive range of texture features, while CVN realizes accurate localization by shuffling channels. In a word, different volume normalization techniques implicitly embody distinct task-specific priors.
Furthermore, our ablation study in \cref{Sec43} shows that volume normalization does help to improve average performance.

\subsection{Ablation Study}\label{Sec43}
This subsection discusses the impact of some proposed methods on performance. The study is conducted on full MVTecAD, and other settings are the same as \cref{Sec42}. We ablate four methods from baseline individually. 
In particular, experiment \numroman{1} is based on latent Gaussian assumption and \cref{eq8:probtrans} without volume normalization. For experiment \numroman{2}, the category-independent zero template $\Bar{z}_d=0$ is applied.
Then, Experiment \numroman{3} does not adopt the method of \cref{eq10:anomalious} but composes the pyramid first and localizes its difference later. Finally, experiment \numroman{4} adopts a spatial version of the loss function instead of \cref{eq9:fourierloss}.
The result of the above ablation experiments is shown in \cref{tab2}.

\begin{table}[htbp]
    \centering
    \setlength{\abovecaptionskip}{0.0cm} 
    \setlength{\belowcaptionskip}{-0.3cm} 
    \caption{
        The ablation study on full MVTecAD. 
        For each cell in the table, the first row is \colorbox{green!5}{Pixel-AUROC\%} and the second is \colorbox{blue!5}{AUPRO\%}.
        The number within parentheses means the change relative to baseline, the larger absolute value with larger importance.
    }

    \resizebox{.4\textwidth}{!}{
        
\begin{tabular}{lccc}
\toprule
\multicolumn{1}{c}{\multirow{2}[4]{*}{\textbf{Method}}} & \multicolumn{2}{c}{\textbf{Classes}} & \multirow{2}[4]{*}{\textbf{MEAN}} \\
\cmidrule{2-3} & Texture & Object & \multicolumn{1}{c}{} \\
\midrule

\multirow{2}[2]{*}{Ours (baseline)} 
& \cellcolor{green!5}\textbf{95.2\footnotesize{(+0.0)}} & \cellcolor{green!5}\textbf{95.7\footnotesize{(+0.0)}} & \cellcolor{green!5}\textbf{95.5\footnotesize{(+0.0)}} \\
& \cellcolor{blue!5} \textbf{95.1\footnotesize{(+0.0)}} & \cellcolor{blue!5} \textbf{93.5\footnotesize{(+0.0)}} & \cellcolor{blue!5} \textbf{94.0\footnotesize{(+0.0)}} \\
\midrule

\multirow{2}[2]{*}{\makecell[l]{\enumroman{1} w/o Volume\\ Normalization}} 
& \cellcolor{green!5} 89.4\footnotesize{(-5.8)} & \cellcolor{green!5} 85.2\footnotesize{(-10.5)} & \cellcolor{green!5} 86.6\footnotesize{(-8.9)} \\
& \cellcolor{blue!5}  87.5\footnotesize{(-7.6)} & \cellcolor{blue!5}  83.6\footnotesize{(-9.9)}  & \cellcolor{blue!5} 84.9\footnotesize{(-9.1)} \\
\midrule

\multirow{2}[2]{*}{\makecell[l]{\enumroman{2} w/o Latent\\ Template}} 
& \cellcolor{green!5} 93.1\footnotesize{(-2.1)} & \cellcolor{green!5} 90.7\footnotesize{(-5.0)} & \cellcolor{green!5} 91.5\footnotesize{(-4.0)} \\
& \cellcolor{blue!5}  91.9\footnotesize{(-3.2)} & \cellcolor{blue!5}  84.7\footnotesize{(-8.8)} & \cellcolor{blue!5}  87.1\footnotesize{(-6.9)} \\
\midrule

\multirow{2}[2]{*}{\makecell[l]{\enumroman{3} w/o Pyramid\\ Difference}} 
& \cellcolor{green!5} 87.8\footnotesize{(-7.4)} & \cellcolor{green!5} 93.1\footnotesize{(-2.6)} & \cellcolor{green!5} 91.3\footnotesize{(-4.2)} \\
& \cellcolor{blue!5}   87.8\footnotesize{(-7.3)} & \cellcolor{blue!5} 89.4\footnotesize{(-4.1)} & \cellcolor{blue!5}  88.9\footnotesize{(-5.1)} \\
\midrule

\multirow{2}[2]{*}{\makecell[l]{\enumroman{4} w/o Fourier\\ Loss}} 
& \cellcolor{green!5} 92.0\footnotesize{(-3.2)} & \cellcolor{green!5} 93.3\footnotesize{(-2.4)} & \cellcolor{green!5} 92.9\footnotesize{(-2.6)} \\
& \cellcolor{blue!5}  92.8\footnotesize{(-2.3)} & \cellcolor{blue!5}  91.9\footnotesize{(-1.6)} & \cellcolor{blue!5}  92.2\footnotesize{(-1.8)} \\
\bottomrule

\end{tabular}

    }
    \label{tab2}%
\end{table}%

\cref{tab2} demonstrates that experiments \numroman{1}-\numroman{4} present various performance degradation.
Experiment \numroman{1} has the largest average degradation, with the object classes being more affected. Although the non-volume-preserving enables larger outputs and higher Image-AUROC performance, the implicit prior in volume normalization discussed in \cref{Sec42} is more helpful for generalization.
For experiment \numroman{2}, it shows that the latent template benefits the performance, and object classes are improved greatly. It is because the category-specific latent template reduces intra-class variance, helping convergence during training.
Then, experiment \numroman{3} suggests that multi-scale differences had a more pronounced impact on textural classes, as higher-level operators with larger receptive fields correspond to large defects.
Finally, Experiment \numroman{4} reveals that Fourier loss (\ref{eq9:fourierloss}) is the icing on the cake that helps performance improvement. 
To summarize, methods \numroman{1}-\numroman{3} are critical for the proposed model, while tricks \numroman{4} help further improvements.

\subsection{Anomaly Localization}\label{Sec44}
\begin{table*}[htbp]
    \centering
    \setlength{\abovecaptionskip}{0.0cm} 
    \setlength{\belowcaptionskip}{-0.3cm} 
    \caption{
    Quantitative results of various challenging methods on MVTecAD.
    In the table, the fully normalized flow method is labeled as FNF, while the abbreviations Res18, WRes50, EffiB5, and DTD are denoted as ResNet18, Wide-ResNet50-2, EfficientNet-B5, and Describable Textures Dataset, respectively.
    For each case in the table, the first row is \colorbox{green!5}{Pixel-AUROC\%} and the second is \colorbox{blue!5}{AUPRO\%}, where the best results are marked in bold. 
    }

    \resizebox{\textwidth}{!}{
        
\begin{tabular}{ccccccccccccccc}
\toprule
\multicolumn{1}{c}{\multirow{2}[2]{*}{\textbf{\makecell{External\\Prior}}}} & \multirow{2}[2]{*}{\textbf{Methods}} & \multicolumn{1}{c}{\multirow{2}[2]{*}{\textbf{carpet}}} & \multicolumn{1}{c}{\multirow{2}[2]{*}{\textbf{leather}}} & \multicolumn{1}{c}{\multirow{2}[2]{*}{\textbf{tile}}} & \multicolumn{1}{c}{\multirow{2}[2]{*}{\textbf{wood}}} & \multicolumn{1}{c}{\multirow{2}[2]{*}{\textbf{bottle}}} & \multicolumn{1}{c}{\multirow{2}[2]{*}{\textbf{cable}}} & \multicolumn{1}{c}{\multirow{2}[2]{*}{\textbf{capsule}}} & \multicolumn{1}{c}{\multirow{2}[2]{*}{\textbf{hazelnut}}} & \multicolumn{1}{c}{\multirow{2}[2]{*}{\textbf{pill}}} & \multicolumn{1}{c}{\multirow{2}[2]{*}{\textbf{toothbrush}}} & \multicolumn{1}{c}{\multirow{2}[2]{*}{\textbf{transistor}}} & \multicolumn{1}{c}{\multirow{2}[2]{*}{\textbf{zipper}}} & \multicolumn{1}{c}{\multirow{2}[2]{*}{\textbf{MEAN}}} \\
    & \multicolumn{1}{c}{} &       &       &       &       &       &       &       &       &       &       &       &       &  \\
\midrule
\multicolumn{1}{c}{\multirow{8}[8]{*}{$\times$}}

& \multirow{2}[2]{*}{AnoGAN\cite{AnoGAN}} & 
\cellcolor{green!5} 54.2  & \cellcolor{green!5} 64.1  & \cellcolor{green!5} 49.7  & \cellcolor{green!5} 62.1  & \cellcolor{green!5} 85.8  & \cellcolor{green!5} 78.0    & \cellcolor{green!5} 84.1  & \cellcolor{green!5} 87.1  & \cellcolor{green!5} 86.8  & \cellcolor{green!5} 90.0    & \cellcolor{green!5} 79.9  & \cellcolor{green!5} 78.1  & \cellcolor{green!5} 75.0 \\
& \multicolumn{1}{c}{} & 
\cellcolor{blue!5} 20.4  & \cellcolor{blue!5} 37.8  & \cellcolor{blue!5} 17.7  & \cellcolor{blue!5} 38.6  & \cellcolor{blue!5} 62.0    & \cellcolor{blue!5} 38.3  & \cellcolor{blue!5} 30.6  & \cellcolor{blue!5} 69.8  & \cellcolor{blue!5} 77.6  & \cellcolor{blue!5} 74.9  & \cellcolor{blue!5} 54.9  & \cellcolor{blue!5} 46.7  & \cellcolor{blue!5} 47.4 \\
\cmidrule{2-15}          

& \multirow{2}[2]{*}{Vanilla VAE\cite{VanillaVAE}} & 
\cellcolor{green!5} 62.0    & \cellcolor{green!5} 83.5  & \cellcolor{green!5} 52.0    & \cellcolor{green!5} 69.9  & \cellcolor{green!5} 89.4  & \cellcolor{green!5} 81.6  & \cellcolor{green!5} 90.7  & \cellcolor{green!5} 95.1  & \cellcolor{green!5} 87.9  & \cellcolor{green!5} 95.3  & \cellcolor{green!5} 85.1  & \cellcolor{green!5} 77.5  & \cellcolor{green!5} 80.8 \\
& \multicolumn{1}{c}{} & 
\cellcolor{blue!5} 61.9  & \cellcolor{blue!5} 64.9  & \cellcolor{blue!5} 24.2  & \cellcolor{blue!5} 57.8  & \cellcolor{blue!5} 70.5  & \cellcolor{blue!5} 77.9  & \cellcolor{blue!5} 77.9  & \cellcolor{blue!5} 77.0    & \cellcolor{blue!5} 79.3  & \cellcolor{blue!5} 85.4  & \cellcolor{blue!5} 61.0    & \cellcolor{blue!5} 60.8  & \cellcolor{blue!5} 66.6 \\
\cmidrule{2-15}          

& \multirow{2}[2]{*}{AE-SSIM\cite{AE-SSIM}} & 
\cellcolor{green!5} 87.0    & \cellcolor{green!5} 78.0    & \cellcolor{green!5} 59.0    & \cellcolor{green!5} 73.0    & \cellcolor{green!5} 93.0    & \cellcolor{green!5} 82.0    & \cellcolor{green!5} 94.0    & \cellcolor{green!5} 97.0    & \cellcolor{green!5} 91.0    & \cellcolor{green!5} 92.0    & \cellcolor{green!5} 80.0    & \cellcolor{green!5} 88.0    & \cellcolor{green!5} 84.5 \\
& \multicolumn{1}{c}{} & 
\cellcolor{blue!5} 64.7  & \cellcolor{blue!5} 56.1  & \cellcolor{blue!5} 17.5  & \cellcolor{blue!5} 60.5  & \cellcolor{blue!5} 83.4  & \cellcolor{blue!5} 47.8  & \cellcolor{blue!5} 86.0    & \cellcolor{blue!5} 91.6  & \cellcolor{blue!5} 83.0    & \cellcolor{blue!5} 78.4  & \cellcolor{blue!5} 72.4  & \cellcolor{blue!5} 66.5  & \cellcolor{blue!5} 67.3 \\
\cmidrule{2-15} \multicolumn{1}{c}{\multirow{2}[2]{*}{}} 

& Ours  & 
\cellcolor{green!5} \textbf{90.8}  & \cellcolor{green!5} \textbf{99.6} & \cellcolor{green!5} \textbf{97.9} & \cellcolor{green!5} \textbf{93.8}  & \cellcolor{green!5} \textbf{95.9}  & \cellcolor{green!5} \textbf{92.1}  & \cellcolor{green!5} \textbf{96.1}  & \cellcolor{green!5} \textbf{98.0} & \cellcolor{green!5} \textbf{96.2}  & \cellcolor{green!5} \textbf{98.9} & \cellcolor{green!5} \textbf{97.4}  &  \cellcolor{green!5} \textbf{95.4}  & \cellcolor{green!5} \textbf{96.0} \\
& (FNF) & 
\cellcolor{blue!5} \textbf{91.0}    & \cellcolor{blue!5} \textbf{99.7} & \cellcolor{blue!5} \textbf{95.8}  & \cellcolor{blue!5} \textbf{96.2}  & \cellcolor{blue!5} \textbf{94.0} & \cellcolor{blue!5} \textbf{86.4}  & \cellcolor{blue!5} \textbf{93.1}  & \cellcolor{blue!5} \textbf{97.3}  & \cellcolor{blue!5} \textbf{96.3} & \cellcolor{blue!5} \textbf{97.7}  & \cellcolor{blue!5} \textbf{91.4}  & \cellcolor{blue!5} \textbf{95.1}  & \cellcolor{blue!5} \textbf{94.5} \\
\midrule

\multirow{2}[2]{*}{Res18} & \multirow{2}[2]{*}{S-T\cite{S-T}} & 
\cellcolor{green!5} 93.5  & \cellcolor{green!5} 97.8  & \cellcolor{green!5} 92.5  & \cellcolor{green!5} 92.1  & \cellcolor{green!5} 97.8  & \cellcolor{green!5} 91.9  & \cellcolor{green!5} 96.8  & \cellcolor{green!5} 98.2  & \cellcolor{green!5} \textbf{96.5}  & \cellcolor{green!5} 97.9  & \cellcolor{green!5} 73.7  & \cellcolor{green!5} 95.6  & \cellcolor{green!5} 93.7 \\
& \multicolumn{1}{c}{} & 
\cellcolor{blue!5}  87.9  & \cellcolor{blue!5} 94.5  & \cellcolor{blue!5} 94.6  & \cellcolor{blue!5} 91.1  & \cellcolor{blue!5} 93.1  & \cellcolor{blue!5} 81.8  & \cellcolor{blue!5} 96.8  & \cellcolor{blue!5} 96.5  & \cellcolor{blue!5} \textbf{96.1}  & \cellcolor{blue!5} 93.3  & \cellcolor{blue!5} 66.6  & \cellcolor{blue!5} 95.1  & \cellcolor{blue!5} 90.6 \\
\cmidrule{2-15}      

\multirow{2}[2]{*}{WRes50}    & \multirow{2}[2]{*}{SPADE\cite{SPADE}} & 
\cellcolor{green!5} 97.5  & \cellcolor{green!5} 97.6  & \cellcolor{green!5} 87.4  & \cellcolor{green!5} 88.5  & \cellcolor{green!5} \textbf{98.4} & \cellcolor{green!5} \textbf{97.2} & \cellcolor{green!5} \textbf{99.0} & \cellcolor{green!5} \textbf{99.1} & \cellcolor{green!5} \textbf{96.5} & \cellcolor{green!5} 97.9  & \cellcolor{green!5} 94.1  & \cellcolor{green!5} 96.5  & \cellcolor{green!5} 95.8 \\
& \multicolumn{1}{c}{} & 
\cellcolor{blue!5} 94.7  & \cellcolor{blue!5} 97.2  & \cellcolor{blue!5} 75.9  & \cellcolor{blue!5} 87.4  & \cellcolor{blue!5} 95.5  & \cellcolor{blue!5} 90.9 & \cellcolor{blue!5} 93.7  & \cellcolor{blue!5} 95.4  & \cellcolor{blue!5} 94.6  & \cellcolor{blue!5} 93.5  & \cellcolor{blue!5} \textbf{97.4} & \cellcolor{blue!5} 92.6  & \cellcolor{blue!5} 92.4 \\
\cmidrule{2-15}     

\multirow{2}[2]{*}{WRes50}    & \multirow{2}[2]{*}{PaDiM\cite{PaDiM}} & 
\cellcolor{green!5} \textbf{99.1} & \cellcolor{green!5} 99.2  & \cellcolor{green!5} 94.1  & \cellcolor{green!5} 94.9  & \cellcolor{green!5} 98.3  & \cellcolor{green!5} 96.7  & \cellcolor{green!5} 98.5  & \cellcolor{green!5} 98.2  & \cellcolor{green!5} 95.7  & \cellcolor{green!5} \textbf{98.8}  & \cellcolor{green!5} \textbf{97.5} & \cellcolor{green!5} \textbf{98.5} & \cellcolor{green!5} \textbf{97.5} \\
& \multicolumn{1}{c}{} & 
\cellcolor{blue!5} 96.2  & \cellcolor{blue!5} 97.8  & \cellcolor{blue!5} 86.0    & \cellcolor{blue!5} 91.1  & \cellcolor{blue!5} 94.8  & \cellcolor{blue!5} 88.8  & \cellcolor{blue!5} 93.5  & \cellcolor{blue!5} 92.6  & \cellcolor{blue!5} 92.7  & \cellcolor{blue!5} 93.1  & \cellcolor{blue!5} 84.5  & \cellcolor{blue!5} \textbf{95.9} & 92.3 \\
\cmidrule{2-15}     

\multirow{2}[2]{*}{EffiB5}    & \multirow{2}[2]{*}{CS-Flow\cite{CSFlow}} & 
\cellcolor{green!5} 98.0    & \cellcolor{green!5} 98.4    & \cellcolor{green!5} 93.9    & \cellcolor{green!5} 88.6    & \cellcolor{green!5} 90.9    & \cellcolor{green!5} 95.3    & \cellcolor{green!5} 97.9    & \cellcolor{green!5} 96.3    & \cellcolor{green!5} 95.7    & \cellcolor{green!5} 96.3    & \cellcolor{green!5} 95.5    & \cellcolor{green!5} 96.4    & \cellcolor{green!5} 95.3 \\
& \multicolumn{1}{c}{} & 
\cellcolor{blue!5} \textbf{98.0}  & \cellcolor{blue!5} 98.5  & \cellcolor{blue!5} 94.5  & \cellcolor{blue!5} 92.9  & \cellcolor{blue!5} 88.7  & \cellcolor{blue!5} \textbf{94.0}  & \cellcolor{blue!5} 96.1    & \cellcolor{blue!5} 95.1  & \cellcolor{blue!5} 91.1    & \cellcolor{blue!5} 89.9  & \cellcolor{blue!5} 96.9  & \cellcolor{blue!5} 95.4  & \cellcolor{blue!5} 94.2 \\
\cmidrule{2-15}     

\multirow{2}[2]{*}{DTD}     & \multirow{2}[2]{*}{DRÆM\cite{DRAEM}} & 
\cellcolor{green!5} 94.9 & \cellcolor{green!5} \textbf{96.6}  & \cellcolor{green!5} \textbf{99.6}  & \cellcolor{green!5} \textbf{97.3}  & \cellcolor{green!5} 97.6  & \cellcolor{green!5} 95.4  & \cellcolor{green!5} 94.0  & \cellcolor{green!5} 99.2  & \cellcolor{green!5} 95.0  & \cellcolor{green!5} 98.1  & \cellcolor{green!5} 90.0 & \cellcolor{green!5} 94.4 & \cellcolor{green!5} 96.0 \\
& \multicolumn{1}{c}{} & 
\cellcolor{blue!5} 96.1  & \cellcolor{blue!5} 97.9  & \cellcolor{blue!5} \textbf{99.7}  & \cellcolor{blue!5} \textbf{97.9}  & \cellcolor{blue!5} \textbf{97.2}  & \cellcolor{blue!5} 90.4  & \cellcolor{blue!5} 96.5  & \cellcolor{blue!5} \textbf{98.7}  & \cellcolor{blue!5} 93.7  & \cellcolor{blue!5} 97.1  & \cellcolor{blue!5} 92.9  & \cellcolor{blue!5} 94.7 & \cellcolor{blue!5} 96.1 \\
\cmidrule{2-15}    

\multirow{2}[2]{*}{Res18}     & \multirow{2}[2]{*}{Ours}   & 
\cellcolor{green!5} 97.4  & \cellcolor{green!5} 98.7  & \cellcolor{green!5} 97.1  & \cellcolor{green!5} 97.0 & \cellcolor{green!5} 97.8  & \cellcolor{green!5} 91.8  & \cellcolor{green!5} 98.6  & \cellcolor{green!5} 98.1  & \cellcolor{green!5} 96.1  & \cellcolor{green!5} 98.5  & \cellcolor{green!5} 96.9  & \cellcolor{green!5} 96.6  & \cellcolor{green!5} 97.1 \\
& \multicolumn{1}{c}{} & 
\cellcolor{blue!5} 97.2  & \cellcolor{blue!5} \textbf{99.2}  & \cellcolor{blue!5} 97.2  & \cellcolor{blue!5} \textbf{97.9} & \cellcolor{blue!5} 95.5 & \cellcolor{blue!5} 90.3  & \cellcolor{blue!5} \textbf{98.3}  & \cellcolor{blue!5} 98.1 & \cellcolor{blue!5} \textbf{96.1}  & \cellcolor{blue!5} \textbf{97.9} & \cellcolor{blue!5} 94.7  & \cellcolor{blue!5} 95.4  & \cellcolor{blue!5} \textbf{96.5} \\
\bottomrule

\end{tabular}%

    }

    \label{tab3}%
\end{table*}%

\noindent\textbf{MVTecAD.}
We performed defect localization for 12 registered classes in MVTecAD.
In our comparisons, the method based on pre-trained models or using external datasets is viewed as requiring external prior, corresponding to the first column of \cref{tab3}.
In our implementation, we augment textural classes with flips and rotations, each with a probability of 0.5, while object categories do not undergo any augmentation operation.
It is worth noting that those base on complex augmentation or weak supervision is not considered in our comparisons, as our approach is capable of incorporating these techniques to improve performance. The detailed results are shown in \cref{tab3}.

First, we take three methods based on image contrast, AnoGAN\cite{AnoGAN}, Vanilla VAE\cite{VanillaVAE}, and AE-SSIM\cite{AE-SSIM}. They are not dependent on external datasets, so it is fair to compare them with our FNF model.
Furthermore, we also compared our method to those that utilize external priors, such as S-T, SPADE, \etc. All methods were reproduced based on the official implementation or AnomaLib\cite{Anomalib}.
For fair comparisons, we adapted the $1\times 1$ convolution $\mathbf{W}$ to the pre-trained encoder, where the pre-trained encoder is the first two layers of ResNet18 for extracting the image into features of original 1/4 size and with 64 channels.

\begin{figure*}
    \centering
    \setlength{\abovecaptionskip}{0.cm} 
    \setlength{\belowcaptionskip}{-0.6cm} 
    
    \includegraphics[width=\linewidth]{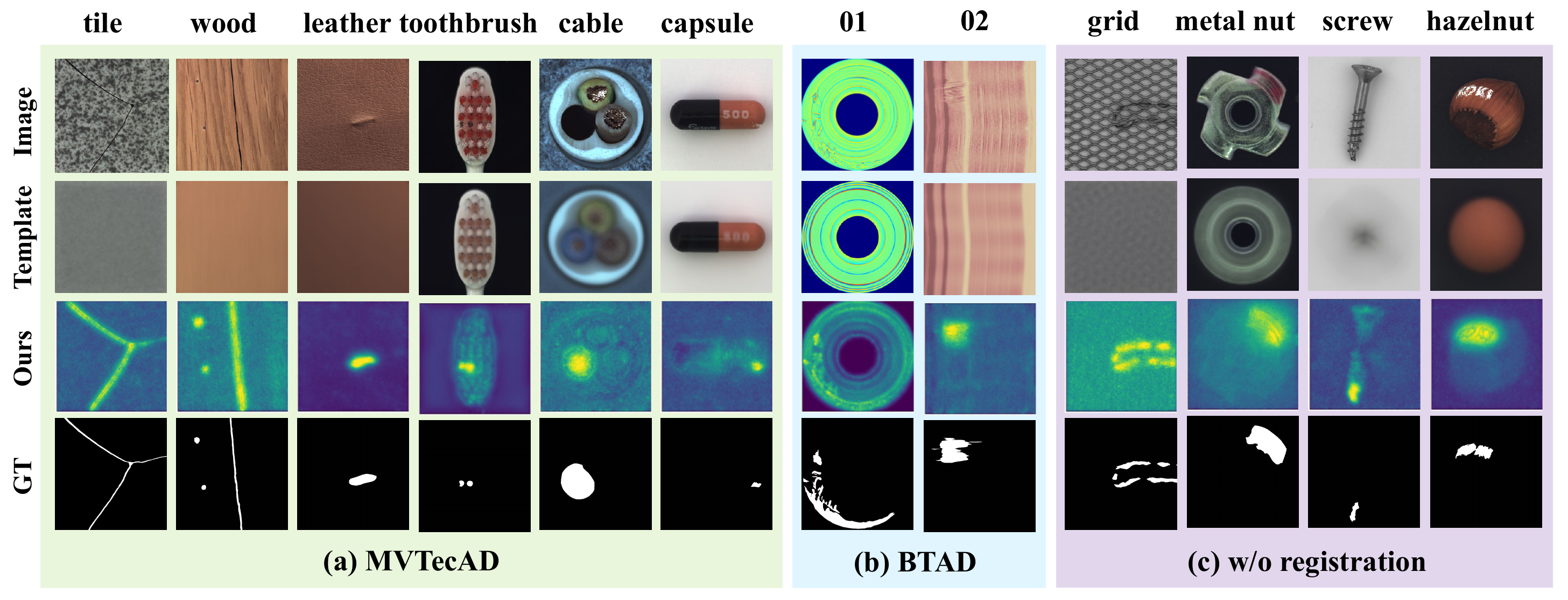}
    \caption{
        Visualization of our results on MVTecAD and BTAD. 
        From top to bottom are original images, estimated image templates, our localization results, and ground truths.
        \textbf{(a)} The six challenging results on MVTecAD. \textbf{(b)} Two representative results on BTAD. \textbf{(c)} Results for the four unregistered categories on MVTecAD.
    }
    \label{fig7}
\end{figure*}

As shown in \cref{tab3}, our FNF method greatly outperforms the comparable methods without external priors, even exceeding S-T, SPADE, and CS-Flow that using external priors.
Most of the reconstruction-based methods in \cref{tab3} suffer from ill-posedness in complex scenarios (\eg, tile and wood.), while our method achieves the best AUPRO score owing to high-resolution contrast in latent space.
However, a larger resolution implies larger intra-class variance, which degrades the overall AUROC performance for hard-to-determine anomaly boundaries.
Furthermore, \cref{fig7}(a) visualizes representative examples of MVTecAD anomaly localization, which shows that our method achieves precise localization with reasonable scale.

\noindent\textbf{BTAD.}
To fully illustrate our superiority, we experimented on the more challenging BTAD dataset without any data augmentation, and other settings are the same as MVTecAD. The detailed result in \cref{tab4} shows that our method also achieves state-of-the-art performance, and \cref{fig7}(b) visualizes representative examples of BTAD anomaly localization.

\begin{table}[htbp]
    \centering
    \setlength{\abovecaptionskip}{0.0cm} 
    \setlength{\belowcaptionskip}{-0.1cm} 
    \caption{
        Quantitative results of various challenging methods on BTAD.
        For each case in the table, the first row is \colorbox{red!5}{Image-AUROC\%} and the second is \colorbox{green!5}{Pixel-AUROC\%}, where the best results are marked in bold.
    }

    \resizebox{.35\textwidth}{!}{

\begin{tabular}{ccccc}
\toprule
\multicolumn{1}{c}{\multirow{2}[4]{*}{\textbf{Methods}}} & \multicolumn{3}{c}{\textbf{Classes}} & \multicolumn{1}{c}{\multirow{2}[4]{*}{\textbf{MEAM}}} \\
\cmidrule{2-4}          & 01     & 02     & 03     &  \\
\midrule

\multicolumn{1}{c}{\multirow{2}[2]{*}{VT-ADL\cite{VT-ADL}}} 
& \cellcolor{red!5} 97.6  & \cellcolor{red!5} 71.0  & \cellcolor{red!5} 82.6  & \cellcolor{red!5} 83.7  \\
& \cellcolor{green!5} \textbf{99.0 } & \cellcolor{green!5} 94.0  & \cellcolor{green!5} 77.0  & \cellcolor{green!5} 90.0  \\
\midrule

\multicolumn{1}{c}{\multirow{2}[2]{*}{P-SVDD\cite{PSVDD}}} 
& \cellcolor{red!5} 95.7  & \cellcolor{red!5} 72.1  & \cellcolor{red!5} 82.1  & \cellcolor{red!5} 83.3  \\
& \cellcolor{green!5} 91.6  & \cellcolor{green!5} 93.6  & \cellcolor{green!5} 91.0  & \cellcolor{green!5} 92.1  \\
\midrule

\multicolumn{1}{c}{\multirow{2}[2]{*}{SPADE\cite{SPADE}}} 
& \cellcolor{red!5} 91.4  & \cellcolor{red!5} 71.4  & \cellcolor{red!5} \textbf{99.9 } & \cellcolor{red!5} 87.6  \\
& \cellcolor{green!5} 97.3  & \cellcolor{green!5} 94.4  & \cellcolor{green!5} 99.1  & \cellcolor{green!5} 96.9  \\
\midrule

\multicolumn{1}{c}{\multirow{2}[2]{*}{PatchCore\cite{PatchCore}}} 
& \cellcolor{red!5} 90.9  & \cellcolor{red!5} 79.3  & \cellcolor{red!5} 99.8  & \cellcolor{red!5} 90.0  \\
& \cellcolor{green!5} 95.5  & \cellcolor{green!5} 94.7  & \cellcolor{green!5} \textbf{99.3 } & \cellcolor{green!5} 96.5  \\
\midrule

\multicolumn{1}{c}{\multirow{2}[2]{*}{PaDiM\cite{PaDiM}}} 
& \cellcolor{red!5} 99.8  & \cellcolor{red!5} 82.0  & \cellcolor{red!5} 99.4  & \cellcolor{red!5} 93.7  \\
& \cellcolor{green!5} 97.0  & \cellcolor{green!5} 96.0  & \cellcolor{green!5} 98.8  & \cellcolor{green!5} 97.3  \\
\midrule

\multicolumn{1}{c}{\multirow{2}[2]{*}{Ours (Res18)}} 
& \cellcolor{red!5} \textbf{100.0 } & \cellcolor{red!5} \textbf{88.2 } & \cellcolor{red!5} 99.3  & \cellcolor{red!5} \textbf{95.8 } \\
& \cellcolor{green!5} 97.4  & \cellcolor{green!5} \textbf{97.6 } & \cellcolor{green!5} 98.1  & \cellcolor{green!5} \textbf{97.7 } \\
\bottomrule

\end{tabular}%

    }

    \label{tab4}%
\end{table}%

\subsection{Study on unregistered categories}\label{Sec45}
In principle, template-based methods require pixel-level registration between images and templates, which is typically satisfied in most real-world scenarios, but may fail in some cases. This subsection explores the performance of the proposed method on unregistered (\eg rotation, shift) categories (\eg grid, metal nut, screw, and hazelnut), as shown in \cref{tab5}.
The reconstruction-based AE-SSIM heavily relies on pixel-level contrast, which decreases the average localization accuracy (AUPRO\%). In contrast, the anomaly-based SPADE avoids registration issues and achieves better performance. Fortunately, our ResNet18-based method, which utilizes normalizing flow to reduce patch variance, remains competitive in unregistered scenes, although it falls short of state-of-the-art performance. 
We visualize these categories in \cref{fig7}(c).

\begin{table}[htbp]
    \centering
    \setlength{\abovecaptionskip}{0.cm} 
    \setlength{\belowcaptionskip}{-0.2cm} 
    \caption{
        Quantitative results on unregistered categories without Rough Registration (RR) or Fine Registration (FR).
        For each case in the table, the first row is \colorbox{green!5}{Pixel-AUROC\%} and the second is \colorbox{blue!5}{AUPRO\%}, where the best results are marked in bold.
    }

    \resizebox{.4\textwidth}{!}{

\begin{tabular}{cccccc}
\toprule
\multicolumn{1}{c}{\multirow{3}[6]{*}{\textbf{Methods}}} & \multicolumn{4}{c}{\textbf{Classes}} & \multicolumn{1}{c}{\multirow{3}[6]{*}{\textbf{MEAN}}} \\
\cmidrule{2-5}          & \multicolumn{3}{c}{w/o RR} & \multicolumn{1}{c}{w/o FR} &  \\
\cmidrule{2-5}          & \multicolumn{1}{c}{grid} & \multicolumn{1}{c}{metal nut} & \multicolumn{1}{c}{screw} & \multicolumn{1}{c}{hazelnut} &  \\

\midrule    \multicolumn{1}{c}{\multirow{2}[2]{*}{AE-SSIM\cite{AE-SSIM}}} 
& \cellcolor{green!5} 94.0    & \cellcolor{green!5} 89.0    & \cellcolor{green!5} 96.0    & \cellcolor{green!5} 97.0    & \cellcolor{green!5} 94.0 \\
& \cellcolor{blue!5} 84.9  & \cellcolor{blue!5} 60.3  & \cellcolor{blue!5} 88.7  & \cellcolor{blue!5} 91.6  & \cellcolor{blue!5} 81.4 \\
\midrule

\multicolumn{1}{c}{\multirow{2}[2]{*}{SPADE\cite{SPADE}}} 
& \cellcolor{green!5} 93.7  & \cellcolor{green!5} \textbf{98.1} & \cellcolor{green!5} \textbf{98.9} & \cellcolor{green!5} \textbf{99.1} & \cellcolor{green!5} \textbf{97.5} \\
& \cellcolor{blue!5} 86.7  & \cellcolor{blue!5} \textbf{94.4} & \cellcolor{blue!5} \textbf{96.0} & \cellcolor{blue!5} 95.4  & \cellcolor{blue!5} 93.1 \\
\midrule

\multicolumn{1}{c}{\multirow{2}[2]{*}{Ours (Res18)}} 
& \cellcolor{green!5} \textbf{95.7} & \cellcolor{green!5} 97.2  & \cellcolor{green!5} 94.6  & \cellcolor{green!5} 98.1  & \cellcolor{green!5} 96.4 \\
& \cellcolor{blue!5} \textbf{94.3} & \cellcolor{blue!5} 91.4  & \cellcolor{blue!5} 94.7  & \cellcolor{blue!5} \textbf{98.1} & \cellcolor{blue!5} \textbf{94.6} \\
\bottomrule

\end{tabular}%

    }

    \label{tab5}%
\end{table}%

\section{Conclusion}

In this paper, we propose PyramidFlow, the first fully normalizing flow method based on the latent template-based contrastive paradigm, utilizing pyramid-like normalizing flows and volume normalization, enabling high-resolution defect contrastive localization. Our method can be trained end-to-end from scratch, similar to UNet, and our comprehensive experiments demonstrate that it outperforms comparable algorithms that do not use external priors, even achieving state-of-the-art performance in complex scenarios. While experiments on unregistered categories show that our method falls short of state-of-the-art, it still exhibits competitive performance. Future research will focus on improving performance in such scenarios.

\vspace{1em}
\noindent \textbf{Acknowledgment.} 
We would like to extend sincere appreciation to \textit{Jiabao Lei} for his valuable guidance and insightful suggestions, which greatly contributed to the success of this work.


{\small
\bibliographystyle{ieee_fullname}
\bibliography{egbib}
}

\end{document}


\title{Supplementary Material for \\ PyramidFlow: High-Resolution Defect Contrastive Localization using Pyramid Normalizing Flow
}

\maketitle

\thispagestyle{empty} 

\section{Implementation Details}

\subsection{Experimental Settings}
\noindent\textbf{Hardware. } We implemented our models in Python3.8 and Pytorch1.10. Experiments are run on NVIDIA GTX3060 GPUs.

\noindent\textbf{Baseline method. } We train our baseline model on $256\times 256$ image. During all experiments, the training batch size is fixed to 2. Model parameters are updated using Adam optimizer with a constant learning rate of $2\times 10^{-4}$, epsilon of $1\times 10^{-4}$, weight decay of $1\times 10^{-5}$, and beta parameters of $(0.5,0.9)$. In addition, we apply gradient clipping with a maximum gradient of $1.0$ for training stability.

\noindent\textbf{Pre-trained method. } For the pre-trained version of PyramidFlow, we used ImageNet-pretrained ResNet18 from torchvision. The pre-trained encoder is the first two layers of ResNet18 for extracting the features from $1024\times 1024$ image to $256\times 256$ features with 64 channels.

\subsection{Model Architecture}
In this subsection, we provide the detailed architecture of the proposed PyramidFlow, including invertible pyramids, pyramid coupling blocks, and volume normalization.

\noindent\textbf{Invertible Pyramid. }
The invertible pyramid is inspired by the Laplacian pyramid, which is commonly used in image processing. In invertible pyramids, the pyramid decomposition and composition are performed on  the per-channel features. The linear downsampling operator $D(\cdot)$ first applies a Gaussian filter with kernel size $5\times 5$, then downsamples using nearest-neighbor interpolation. In contrast, upsampling $U(\cdot)$ performs nearest-neighbor interpolation before applying Gaussian filtering.

\noindent\textbf{Pyramid Coupling Block. }%
For the example of dual coupling blocks, denoting the feature notations as shown in Fig. 3(d), the corresponding pseudocode is described in \cref{alg1}. It is mainly composed of three custom functions - \textit{AffineParamBlock, VolumeNorm2d, }and \textit{InvConv}.

\noindent\textbf{Volume Normalization. }%
The proposed volume normalization is similar to some normalization techniques such as Batch Normalization, but without normalizing the standard deviation. Taking Channel Volume Normalization (CVN) as an example, it can be described by the \cref{alg2}.

\section{More Experiment Results}

\subsection{Detailed Ablation Results}
We present the detailed ablation results of Sec 4.3, as shown in \cref{tab:abl,tab:ab2}.

\noindent\textbf{Textural Image. }
As shown in \cref{tab:abl}. For most textural categories, occurring performance degradation when the proposed methods are ablated.
However, the results on the carpet show abnormal performance improvement.
This means that inductive bias brings positive or negative effects on various categories.

\noindent\textbf{Object Image. }
As shown in \cref{tab:ab2}. Due to the image patch in object categories with larger variances, the influence of volume normalization and the latent template is also larger.
The performance of the object categories is less influenced by pyramid difference, indicating that multi-scale is not a critical factor for object defect detection.

\begin{center}
\begin{minipage}{0.7\linewidth}
    \begin{algorithm}[H]
    \ttfamily\small
    \caption{Dual Coupling Block. (Python-like Pseudocode)}
        \begin{algorithmic}
        \REQUIRE {$x_0,x_1,x_2$}
        \ENSURE  {$z_0,z_1,z_2$}

        $x_{cat0}$ = Interpolate($x_0$, $x_1$.shape) \\
        $x_{cat2}$ = Interpolate($x_2$, $x_1$.shape)\\
        $x_{cat}$ = Concat($x_{cat0}$, $x_{cat2}$)\\
        $s_1$, $t_1$ = AffineParamBlock($x_{cat}$)\\
        $y_1$ = $\exp{(s_1)}\odot x_1 + t_1$\\
        $z_0$, $z_1$, $z_2$ = $x_0$, InvConv($y_1$), $x_2$ \\
        \vspace{\baselineskip} 
        \textbf{def} AffineParamBlock(x, clamp=2):\\
         \quad params = CNN2d(x) \textcolor[rgb]{0,0.5,0}{\% only two convolutional layers and one activation layer}\\
         \quad $s_0$, $t$ = Chunk2d(params)\\
         \quad $s$ = VolumeNorm2d(clamp*0.636*atan($s_0$/clamp)) \textcolor[rgb]{0,0.5,0}{\% as shown in \cref{alg2}. Where 0.636 is an approximation of $2/\pi$.} \\
         \quad \textbf{return} $s,t$\\
        \vspace{\baselineskip} 
         \textbf{def} InvConv(y):\\
         \quad $\Tilde{s}_i$ = $s_i$ - mean($s_i$)\\
         \quad kernel = $\textbf{PL}(\textbf{U}+\text{diag}(\exp{(\Tilde{s}_i)})$\\
         \quad z = Conv2d(y, kernel)\\
         \quad \textbf{return} z
        
        \end{algorithmic}\label{alg1}
    \end{algorithm}

\end{minipage}
\end{center}

\begin{center}
\begin{minipage}{0.7\linewidth}
    \begin{algorithm}[H]
    \ttfamily\small
    \caption{Volume Normalization. (Pytorch-like Pseudocode)}
        \begin{algorithmic}
        \REQUIRE {input $x$, momentum $\beta$}
        \ENSURE  {output $y$}
        
         \textbf{def} VolumeNorm2d($x$, $\beta=0.1$):\\
         \quad \textbf{if} training:\\
         \quad\quad $\Bar{x}$ = mean($x$, dim=1) \textcolor[rgb]{0,0.5,0}{\% CVN: zero-mean normalization along channel dimensions}\\
         \quad\quad $y$ = $x$ - $\Bar{x}$\\
         \quad\quad $\Bar{x}_{running}=(1-\beta)\times \bar{x}_{running}+ \beta\times\bar{x}$  \textcolor[rgb]{0,0.5,0}{\% update running mean}\\
         \quad \textbf{else:}\\
         \quad\quad $y$ = $x$ - $\Bar{x}_{running}$\\
         \quad \textbf{return } $y$

        \end{algorithmic}\label{alg2}
    \end{algorithm}
    
\end{minipage}
\end{center}

\begin{table}[htbp]
  \centering
    \setlength{\abovecaptionskip}{0.cm} 
    \setlength{\belowcaptionskip}{-0.6cm} 
  \caption{
    The ablation study on textural images in MVTecAD. 
    For each cell in the table, the first row is \colorbox{green!5}{Pixel-AUROC\%} and the second is \colorbox{blue!5}{AUPRO\%}.
  }
  \resizebox{0.55\linewidth}{!}{

\begin{tabular}{lcccccc}
\toprule
\multicolumn{1}{c}{\multirow{2}[4]{*}{\textbf{Method}}} & \multicolumn{5}{c}{\textbf{Texture}}  & \multirow{2}[4]{*}{\textbf{Mean}} \\
\cmidrule{2-6}          & \multicolumn{1}{c}{carpet} & \multicolumn{1}{c}{grid} & \multicolumn{1}{c}{leather} & \multicolumn{1}{c}{tile} & \multicolumn{1}{c}{wood} &  \\
\midrule

\multirow{2}[2]{*}{Ours (baseline)} 
& \cellcolor{green!5} 90.8  & \cellcolor{green!5} 94.2  & \cellcolor{green!5} 99.6  & \cellcolor{green!5} 97.9  & \cellcolor{green!5} 93.8  & \cellcolor{green!5} 95.2  \\
& \cellcolor{blue!5} 91.0  & \cellcolor{blue!5} 92.7  & \cellcolor{blue!5} 99.7  & \cellcolor{blue!5} 95.8  & \cellcolor{blue!5} 96.2  & \cellcolor{blue!5} 95.1  \\
\midrule

\multirow{2}[2]{*}{\makecell[l]{\enumroman{1} w/o Volume\\ Normalization}} 
& \cellcolor{green!5} 93.5  & \cellcolor{green!5} 88.5  & \cellcolor{green!5} 99.5  & \cellcolor{green!5} 74.4  & \cellcolor{green!5} 91.3  & \cellcolor{green!5} 89.4  \\
& \cellcolor{blue!5} 93.7  & \cellcolor{blue!5} 88.1  & \cellcolor{blue!5} 95.5  & \cellcolor{blue!5} 65.7  & \cellcolor{blue!5} 94.2  & \cellcolor{blue!5} 87.5  \\
\midrule

\multirow{2}[2]{*}{\makecell[l]{\enumroman{2} w/o Latent\\ Template}} 
& \cellcolor{green!5} 91.8  & \cellcolor{green!5} 86.8  & \cellcolor{green!5} 99.4  & \cellcolor{green!5} 94.8  & \cellcolor{green!5} 93.0  & \cellcolor{green!5} 93.1  \\
& \cellcolor{blue!5} 91.3  & \cellcolor{blue!5} 88.0  & \cellcolor{blue!5} 97.7  & \cellcolor{blue!5} 89.9  & \cellcolor{blue!5} 92.7  & \cellcolor{blue!5} 91.9  \\
\midrule

\multirow{2}[2]{*}{\makecell[l]{\enumroman{3} w/o Pyramid\\ Difference}} 
& \cellcolor{green!5} 75.9  & \cellcolor{green!5} 78.0  & \cellcolor{green!5} 99.3  & \cellcolor{green!5} 96.0  & \cellcolor{green!5} 89.7  & \cellcolor{green!5} 87.8  \\
& \cellcolor{blue!5} 76.1  & \cellcolor{blue!5} 76.1  & \cellcolor{blue!5} 99.4  & \cellcolor{blue!5} 94.4  & \cellcolor{blue!5} 93.0  & \cellcolor{blue!5} 87.8  \\
\midrule

\multirow{2}[2]{*}{\makecell[l]{\enumroman{4} w/o Fourier\\ Loss}} 
& \cellcolor{green!5} 90.5  & \cellcolor{green!5} 84.3  & \cellcolor{green!5} 99.4  & \cellcolor{green!5} 96.2  & \cellcolor{green!5} 89.7  & \cellcolor{green!5} 92.0  \\
& \cellcolor{blue!5} 91.4  & \cellcolor{blue!5} 86.2  & \cellcolor{blue!5} 99.6  & \cellcolor{blue!5} 92.6  & \cellcolor{blue!5} 94.0  & \cellcolor{blue!5} 92.8  \\
\bottomrule

\end{tabular}%

  }
  \label{tab:abl}%
\end{table}%

\begin{table}[htbp]
  \centering
    \setlength{\abovecaptionskip}{0.cm} 
    \setlength{\belowcaptionskip}{-0.6cm} 
  \caption{
    The ablation study on object images in MVTecAD. 
    For each cell in the table, the first row is \colorbox{green!5}{Pixel-AUROC\%} and the second is \colorbox{blue!5}{AUPRO\%}.
  }
  \resizebox{0.9\linewidth}{!}{
    
\begin{tabular}{cccccccccccc}
\toprule
\multicolumn{1}{c}{\multirow{2}[4]{*}{\textbf{Method}}} & \multicolumn{10}{c}{\textbf{Object}}                                          & \multirow{2}[4]{*}{\textbf{Mean}} \\
\cmidrule{2-11}          & \multicolumn{1}{c}{bottle} & \multicolumn{1}{c}{cable} & \multicolumn{1}{c}{capsule} & \multicolumn{1}{c}{hazelnut} & \multicolumn{1}{c}{metalnut} & \multicolumn{1}{c}{pill} & \multicolumn{1}{c}{screw} & \multicolumn{1}{c}{toothbrush} & \multicolumn{1}{c}{transistor} & \multicolumn{1}{c}{zipper} &  \\
\midrule

\multirow{2}[2]{*}{Ours (baseline)} 
& \cellcolor{green!5} 95.9  & \cellcolor{green!5} 92.1  & \cellcolor{green!5} 96.1  & \cellcolor{green!5} 98.0  & \cellcolor{green!5} 92.8  & \cellcolor{green!5} 96.2  & \cellcolor{green!5} 94.0  & \cellcolor{green!5} 98.9  & \cellcolor{green!5} 97.4  & \cellcolor{green!5} 95.4  & \cellcolor{green!5} 95.7  \\
& \cellcolor{blue!5} 94.0  & \cellcolor{blue!5} 86.4  & \cellcolor{blue!5} 93.1  & \cellcolor{blue!5} 97.3  & \cellcolor{blue!5} 89.5  & \cellcolor{blue!5} 96.3  & \cellcolor{blue!5} 94.1  & \cellcolor{blue!5} 97.9  & \cellcolor{blue!5} 91.4  & \cellcolor{blue!5} 95.1  & \cellcolor{blue!5} 93.5  \\
\midrule

\multirow{2}[2]{*}{\makecell[l]{\enumroman{1} w/o Volume\\ Normalization}} 
& \cellcolor{green!5} 76.5  & \cellcolor{green!5} 84.7  & \cellcolor{green!5} 82.9  & \cellcolor{green!5} 97.9  & \cellcolor{green!5} 87.9  & \cellcolor{green!5} 94.8  & \cellcolor{green!5} 94.1  & \cellcolor{green!5} 56.4  & \cellcolor{green!5} 82.2  & \cellcolor{green!5} 95.0  & \cellcolor{green!5} 85.2  \\
& \cellcolor{blue!5} 77.8  & \cellcolor{blue!5} 75.1  & \cellcolor{blue!5} 81.3  & \cellcolor{blue!5} 95.4  & \cellcolor{blue!5} 81.5  & \cellcolor{blue!5} 81.5  & \cellcolor{blue!5} 94.0  & \cellcolor{blue!5} 74.2  & \cellcolor{blue!5} 82.7  & \cellcolor{blue!5} 92.6  & \cellcolor{blue!5} 83.6  \\
\midrule

\multirow{2}[2]{*}{\makecell[l]{\enumroman{2} w/o Latent\\ Template}} 
& \cellcolor{green!5} 83.2  & \cellcolor{green!5} 87.8  & \cellcolor{green!5} 90.0  & \cellcolor{green!5} 97.9  & \cellcolor{green!5} 87.6  & \cellcolor{green!5} 94.6  & \cellcolor{green!5} 93.0  & \cellcolor{green!5} 84.7  & \cellcolor{green!5} 94.8  & \cellcolor{green!5} 93.7  & \cellcolor{green!5} 90.7  \\
& \cellcolor{blue!5} 82.4  & \cellcolor{blue!5} 76.6  & \cellcolor{blue!5} 87.3  & \cellcolor{blue!5} 83.9  & \cellcolor{blue!5} 74.2  & \cellcolor{blue!5} 89.3  & \cellcolor{blue!5} 92.7  & \cellcolor{blue!5} 90.7  & \cellcolor{blue!5} 77.4  & \cellcolor{blue!5} 92.8  & \cellcolor{blue!5} 84.7  \\
\midrule

\multirow{2}[2]{*}{\makecell[l]{\enumroman{3} w/o Pyramid\\ Difference}} 
& \cellcolor{green!5} 92.8  & \cellcolor{green!5} 91.4  & \cellcolor{green!5} 96.0  & \cellcolor{green!5} 97.5  & \cellcolor{green!5} 86.4  & \cellcolor{green!5} 95.3  & \cellcolor{green!5} 92.7  & \cellcolor{green!5} 98.0  & \cellcolor{green!5} 95.4  & \cellcolor{green!5} 85.2  & \cellcolor{green!5} 93.1  \\
& \cellcolor{blue!5} 83.4  & \cellcolor{blue!5} 84.1  & \cellcolor{blue!5} 94.0  & \cellcolor{blue!5} 97.6  & \cellcolor{blue!5} 81.2  & \cellcolor{blue!5} 95.4  & \cellcolor{blue!5} 93.1  & \cellcolor{blue!5} 97.1  & \cellcolor{blue!5} 90.7  & \cellcolor{blue!5} 77.2  & \cellcolor{blue!5} 89.4  \\
\midrule

\multirow{2}[2]{*}{\makecell[l]{\enumroman{4} w/o Fourier\\ Loss}} 
& \cellcolor{green!5} 88.0  & \cellcolor{green!5} 88.6  & \cellcolor{green!5} 95.1  & \cellcolor{green!5} 97.3  & \cellcolor{green!5} 88.9  & \cellcolor{green!5} 96.2  & \cellcolor{green!5} 94.2  & \cellcolor{green!5} 98.3  & \cellcolor{green!5} 95.1  & \cellcolor{green!5} 90.9  & \cellcolor{green!5} 93.3  \\
& \cellcolor{blue!5}  88.0  & \cellcolor{blue!5} 81.2  & \cellcolor{blue!5} 94.0  & \cellcolor{blue!5} 98.3  & \cellcolor{blue!5} 89.0  & \cellcolor{blue!5} 96.9  & \cellcolor{blue!5} 94.4  & \cellcolor{blue!5} 97.9  & \cellcolor{blue!5} 88.0  & \cellcolor{blue!5} 90.8  & \cellcolor{blue!5} 91.9  \\
\bottomrule

\end{tabular}%

  }
  \label{tab:ab2}%
\end{table}%

\subsection{More Visualization Results}
\begin{figure}
    \centering
    \setlength{\abovecaptionskip}{0.cm} 
    \setlength{\belowcaptionskip}{-0.6cm} 
    
    \includegraphics[width=\linewidth]{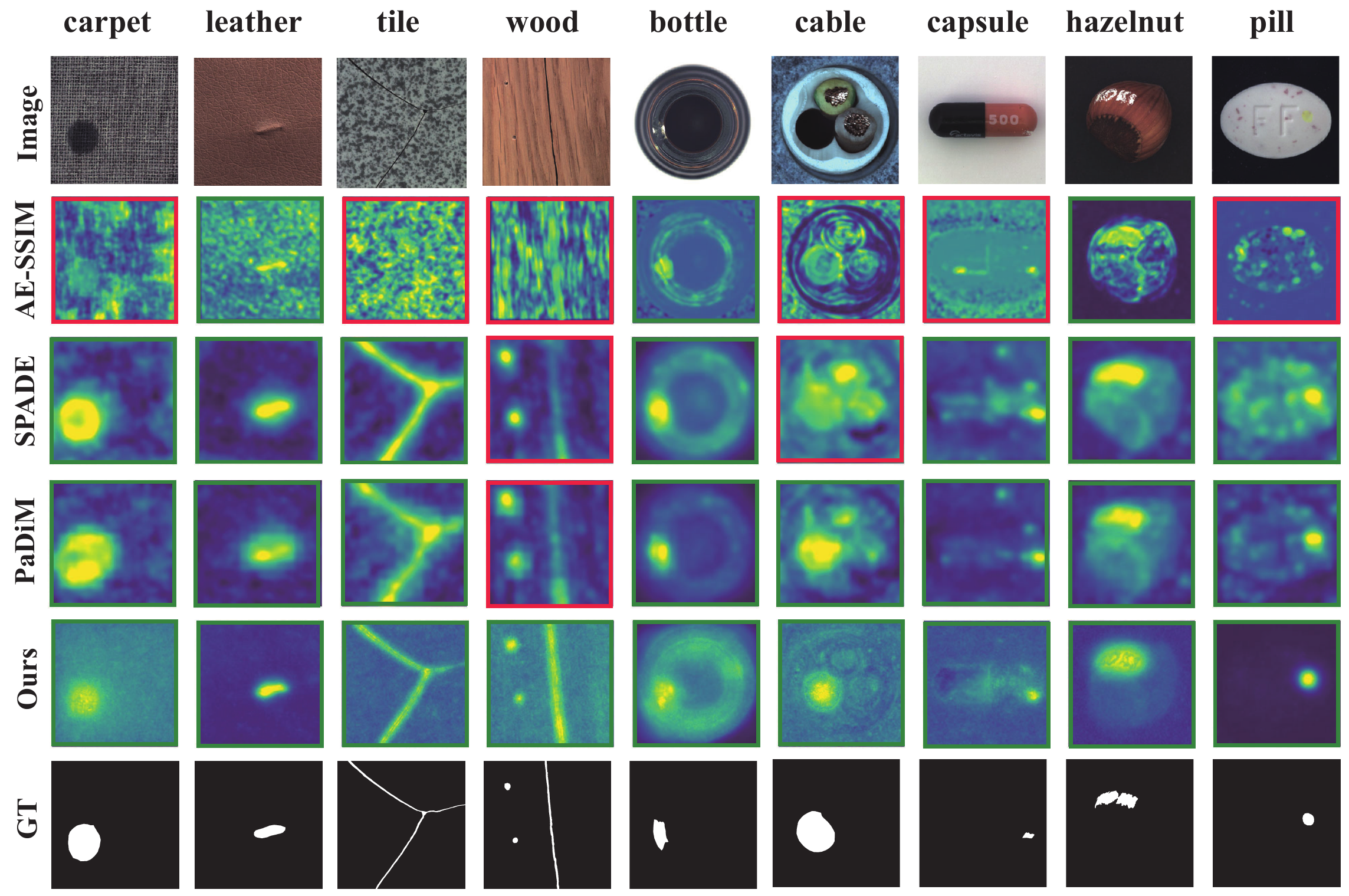}
    
    \caption{
        Visualization of competitive results on MVTecAD. From top to bottom are original images, AE-SSIM results, SPADE results, PaDiM results, our results, and ground truths. 
        The red box indicates the localization is ambiguous and non-unique, while the green indicates successful results.
    }
    \label{fig1}
\end{figure}

\begin{figure}
    \centering
    \setlength{\abovecaptionskip}{0.cm} 
    \setlength{\belowcaptionskip}{-0.6cm} 
    
    \includegraphics[width=\linewidth]{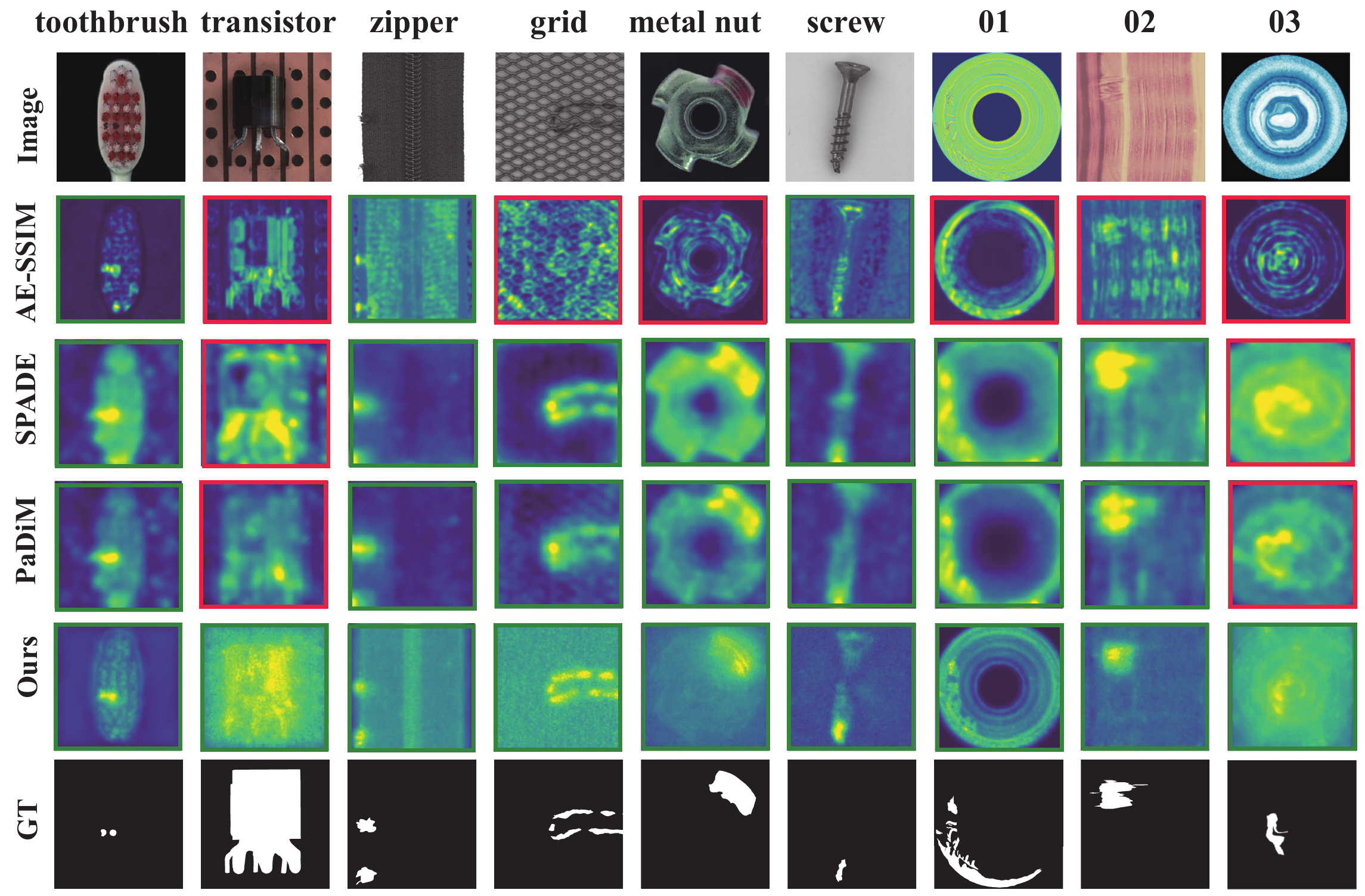}
    
    \caption{
        Visualization of competitive results on MVTecAD and BTAD. From top to bottom are original images, AE-SSIM results, SPADE results, PaDiM results, our results, and ground truths. The last three columns are the results of BTAD.
        The red box indicates the localization is ambiguous and non-unique, while the green indicates successful results.
    }
    \label{fig2}
\end{figure}

In this subsection, we present more visualization results of Sec 4.4. Since many categories, we separated results into two charts for visualization, as shown in \cref{fig1,fig2}.

\noindent\textbf{MVTecAD. }
As \cref{fig1,fig2} shows, AE-SSIM performs better for simple categories, such as the bottle and zipper. However, it does not work in complex scenarios, \eg, it cannot localize carpet defects with fixed patterns or pill defects with high-frequency noises. It is worth noticing that AE-SSIM is a template-based method, which maintains the resolution during processing, enabling preserve the details in defect localization.

SPADE and PaDiM are pre-trained-based methods. They achieve better results in almost all categories but still maintain some shortcomings. On the one hand, their localization results are blurry and larger than ground truths. On the other hand, they cannot localize tiny defects, such as cracks in the wood.

Our proposed PyramidFlow is based on latent templates, which allows for preserving details effectively, with the ability to detect tiny defects and show their scale. In all categories in MVTecAD, our method achieves the best visual performance.

\noindent\textbf{BTAD. }
BTAD is more challenging than MVTecAD, as shown in the last three columns of \cref{fig2}. The AE-SSIM method almost failed in BTAD without beneficial results. For categories 01 and 02, the localization areas of SPADE and PaDiM are obviously larger than ground truths. For the most challenging category 03, their results are incredibly varied from GT.

Our method provides more accurate results for BTAD defect localization. For the 01 categories, the localization results preserve the original details. Categories 02 and 03 also mostly reflect the essential shape of the defect.

